\documentclass[conference]{IEEEtran}
\usepackage[dvipsnames]{xcolor}
\usepackage{xspace}
\usepackage{times}

\usepackage[numbers]{natbib}
\usepackage{multirow}
\usepackage{booktabs}
\usepackage[bookmarks=true]{hyperref}
\usepackage{graphicx}

\usepackage{amsmath}
\usepackage{array}
\usepackage{booktabs}
\usepackage{multicol,multirow}
\usepackage{setspace}
\usepackage{makecell}
\usepackage{tasks}
\usepackage{pifont}
\usepackage{colortbl}
\usepackage{framed}
\usepackage{titletoc}
\usepackage{color}
\usepackage{caption, subcaption, overpic, textpos}
\usepackage{titlesec}
\usepackage{comment}
\usepackage{amssymb}
\usepackage{pifont}

\definecolor{tianyang}{RGB}{255,0,0}   
\definecolor{caizetao}{RGB}{120,120,0}   
\definecolor{yangyuyin}{RGB}{92, 167, 186}
\definecolor{todo}{RGB}{0,0,255}   
\newcommand{\greencheckmark}{\textcolor{teal}{\ding{51}}}
\newcommand{\redxmark}{\textcolor{red}{\ding{55}}}

\newcommand{\modelname}{PPI\xspace}

\newcommand{\yuyin}[1]{{\color{yangyuyin} #1}}

\definecolor{baselinecolor}{gray}{.9}
\newcommand{\baseline}[1]{\cellcolor{baselinecolor}{#1}}
\newcommand{\authorrefmarkl}[1]{\textsuperscript{#1}}

\begin{document}

\title{Gripper Keypose and Object Pointflow as Interfaces for Bimanual Robotic Manipulation}


\author{
  \authorblockN{
    Yuyin Yang\authorrefmarkl{*,1,2}\hspace{1.5em}
    Zetao Cai\authorrefmarkl{*,1,3}\hspace{1.5em}
    Yang Tian\authorrefmarkl{1,4}\hspace{1.5em}
    Jia Zeng\authorrefmarkl{1}\hspace{1.5em}
    Jiangmiao Pang\authorrefmarkl{1}\authorrefmark{2}
  }
  \authorblockA{
   \authorrefmarkl{1}Shanghai AI Laboratory\hspace{2em}
   \authorrefmarkl{2}Fudan University \hspace{2em}
   \authorrefmarkl{3}Zhejiang University\hspace{2em}
   \authorrefmarkl{4}Peking University \\
  }
  \authorblockA{
    \authorrefmarkl{*}Equal contributions \hspace{2em}\authorrefmark{2}Corresponding author \\
  }
  \vspace{0.25em}
  \authorblockA{
   \href{https://yuyinyang3y.github.io/PPI/}{\textbf{\yuyin{https://yuyinyang3y.github.io/PPI/}}}
  }

}

\noindent
\twocolumn[{%
\renewcommand\twocolumn[1][]{#1}
\maketitle
\vspace{-5mm}
\begin{center}
    \centering
    \captionsetup{type=figure}
    \includegraphics[width=1\textwidth]{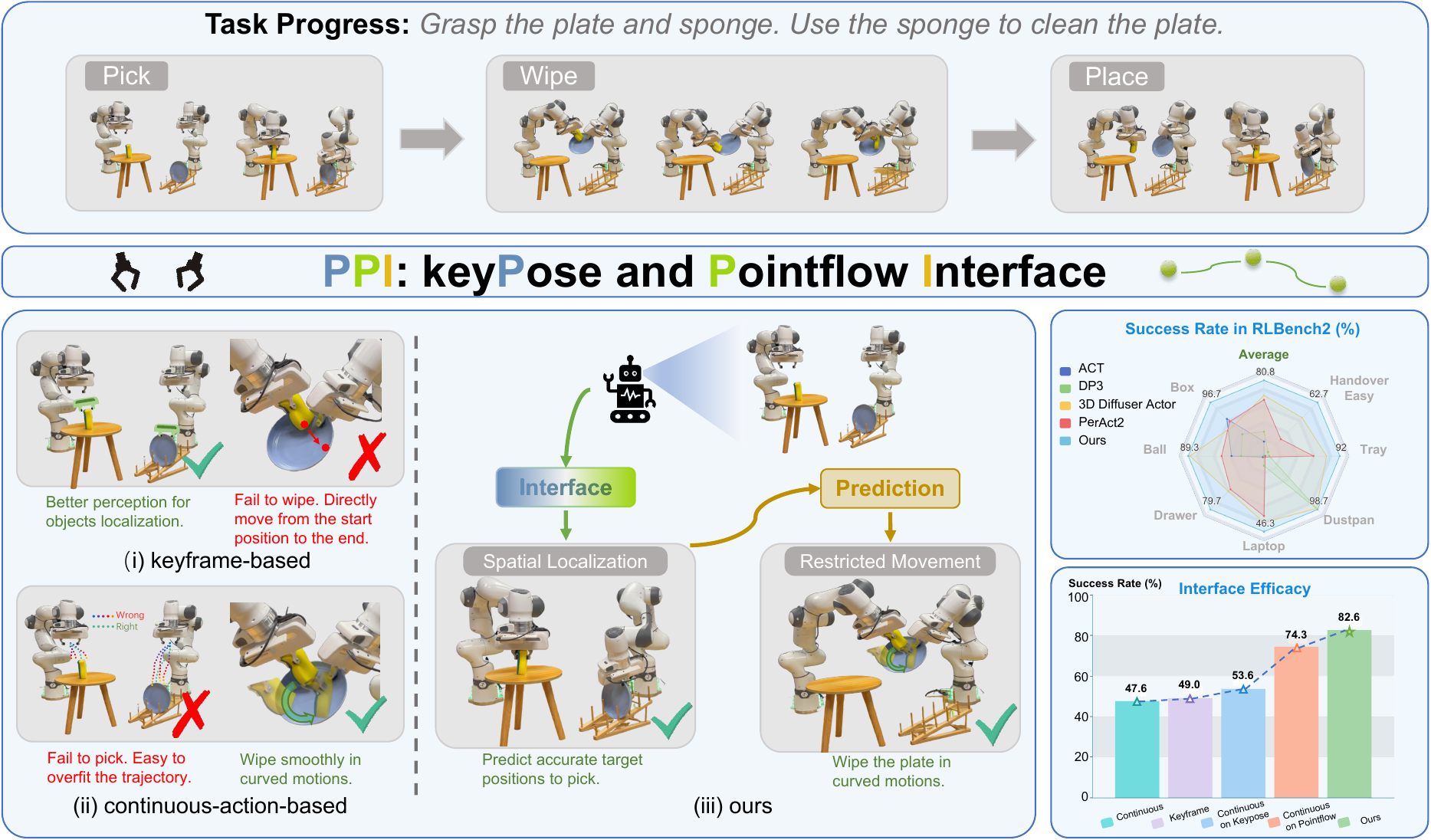}
    \captionof{figure}{{
    In contrast to (i) keyframe-based policies, which excel in spatial localization but struggle with movement restrictions (e.g., curved motion and collision-free actions), and (ii) continuous-action-based policies, which accommodate diverse trajectories but lack strong perception, we introduce a continuous action policy that incorporates two interfaces: target gripper poses and object pointflow,  balancing task diversity with spatial awareness.
    Our model, \modelname, surpasses previous states of the art and consistently outperforms its ablated variants.
    } 
    \label{fig:teaser}
    }
\end{center}%
}]

\begin{abstract}
Bimanual manipulation is a challenging yet crucial robotic capability, demanding precise spatial localization and versatile motion trajectories, which pose significant challenges to existing approaches.
Existing approaches fall into two categories: keyframe-based strategies, which predict gripper poses in keyframes and execute them via motion planners, and continuous control methods, which estimate actions sequentially at each timestep.
The keyframe-based method lacks inter-frame supervision, struggling to perform consistently or execute curved motions, while the continuous method suffers from weaker spatial perception. 
To address these issues, this paper introduces an end-to-end framework \modelname (keyPose and Pointflow Interface), which integrates the prediction of target gripper poses and object pointflow with the continuous actions estimation.  
%
%
These interfaces enable the model to effectively attend to the target manipulation area, while the overall framework guides diverse and collision-free trajectories.  
%
By combining interface predictions with continuous actions estimation, \modelname demonstrates superior performance in diverse bimanual manipulation tasks, providing enhanced spatial localization and satisfying flexibility in handling movement restrictions.  
%
In extensive evaluations, \modelname significantly outperforms prior methods in both simulated and real-world experiments, achieving state-of-the-art performance with a +16.1\% improvement on the RLBench2 simulation benchmark and an average of +27.5\% gain across four challenging real-world tasks.
Notably, \modelname exhibits strong stability, high precision, and remarkable generalization capabilities in real-world scenarios.  
%

\end{abstract}


\IEEEpeerreviewmaketitle

\section{Introduction}
Endowing robots with dexterous bimanual skills similar to humans has become a main focus in robotic manipulation~\citep{rt1,palm-e,openvla,seer,grutopia}.
Recent efforts primarily fall into two categories:
one focuses on ``keyframe'', like~\citep{peract2,3d-diffuser-actor}, which predict actions in the reference frames and execute predictions via Inverse Kinematics (IK) solvers and motion planners~\citep{curobo}.
The other emphasizes ``continuous'' and perform naive behavior cloning at each time step.
For example, ACT~\citep{act} and RDT~\citep{rdt} learn RGB-based manipulation policies from multiple cameras, while DP3~\citep{dp3} integrate 3D scene-level representations into a diffusion policy.

However, due to the temporal granularity of action prediction, the keyframe-based methods predicts actions only at few keyframes.
This sparse supervision encourages the model to focus more on local features, enhancing spatial perception.
Nevertheless, for tasks involving restricted movements (e.g., wiping a plate, which requires curved trajectories), keyframes are difficult to define and the motion planner~\citep{curobo,mplib} tends to output near straight-line paths, making such tasks challenging to accomplish.
For continuous-based methods, they are generally applicable to a wide range of tasks.
Whereas, since they rely on naive behavior cloning with dense supervision on actions, the model tends to ``take shortcuts" by overfitting to seen trajectories (e.g., proprioception).
This results in weaker spatial perception capabilities.
Therefore, implementing diverse general bimanual tasks while preserving strong perceptual capabilities remains a critical challenge. 

To this end, this paper presents a simple yet effective end-to-end interface-based continuous policy that integrates the strengths of previous approaches.
As illustrated in Figure~\ref{fig:teaser}, our model predicts continuous actions conditioned on two key interfaces: the target gripper keypose and object pointflow.
These interfaces enable the model to capture fine-grained spatial features and comprehensively model the interaction between the robot and the object.
We implement a diffusion transformer~\citep{3d-diffuser-actor,dp3} to process both interfaces, naming our approach \modelname.
By distilling spatial knowledge from these interfaces, \modelname strikes a balance between handling diverse tasks and maintaining strong perception capabilities.
Leveraging a unidirectional attention within the transformer, \modelname progressively infers actions and is trained in an end-to-end manner.

We conduct extensive experiments on both simulation and real-world benchmarks.
On the bimanual manipulation benchmark RLBench2~\citep{peract2}, our method achieves a 16.1\% higher success rate across seven representative tasks compared to state-of-the-art baselines.
We further provide comprehensive visualizations to validate the effectiveness of the two interfaces.
Additionally, we evaluate our approach on four challenging real-world tasks~\citep{wu2024gello}, demonstrating superior performance in long-horizon task execution, generalization to unseen objects, robustness to lighting variations, and resilience against visual distractions.

Our contributions are summarized as follows:
\begin{itemize}
    \item 
    We present a novel framework that utilizes keyframe information to guide continuous action generation, improving flexibility in addressing movement constraints.
    \item 
    We propose two effective interfaces—target gripper poses and object pointflow to boost spatial localization and generalization.
    \item 
    We provide comprehensive analyses to validate the power of two interfaces.
    We achieve the state-of-the-art performance on a bimanual simulation benchmark and demonstrate strong robustness, effectiveness, and generalization in real-world long-horizon tasks.
\end{itemize}

\section{Related Works}
\subsection{Behavior Cloning in General Bimanual Manipulation Tasks}
%
Current behavior cloning methods for general bimanual manipulation tasks can be broadly classified into two categories.
The first category involves keyframe-based strategies~\citep{peract2,voxact-b,anygrasp}, where keyframe representations~\citep{foundationpose,dualafford,robokeygen} are learned and executed through motion planners.
Approaches such as PerAct2~\citep{peract2} and VoxAct-B~\citep{voxact-b} predict target gripper poses in a reference frame using voxel-based representations.
Additionally, DualAfford~\citep{dualafford} learns collaborative object-centric affordances and applies heuristic policies for execution.
However, these methods rely on rule-based keyframe split and motion planners, which limits their ability to handle tasks that require irregular motion trajectories (e.g., dishwashing) or strict temporal coordination (e.g., tray lifting).
The second category involves continuous control, where actions are estimated sequentially at each time step.
For instance, ACT~\citep{act} uses an action-chunking transformer to predict actions in an end-to-end manner, while RDT~\citep{rdt} employs a diffusion-based transformer, pre-trained on large robot datasets and fine-tuned on self-collected bimanual data.
BiKC~\citep{bikc}, is an RGB-based hierarchical framework consisting of a high-level keypose predictor and a low-level trajectory generator.
Some works also extend single-arm manipulation policies~\citep{dp,dp3} to the bimanual setting.  
In contrast to these continuous control approaches, our method integrates a 3D semantic neural field~\citep{gendp,d3fields}
and predicts pointflow as an additional interface, thereby enhancing spatial localization capabilities.

\subsection{Flow-based Methods in Robotic Manipulation}
Robot manipulation policies have utilized either 2D pixel-level motion~\citep{cotracker,sgtapose,sam2,fabricflownet} or 3D point-level flow~\citep{flowbot++,taxpose,orientpose} for object interaction.
In 2D flow-based approaches, recent pixel-tracking algorithms~\citep{cotracker} estimate motion flows in robotic video data.
Track2Act~\citep{track2act} integrates a residual strategy atop heuristic and flow-based policies, while ATM~\citep{atm} learns a flow-conditioned behavior cloning policy trained on self-collected, in-domain data.
Im2Flow2Act~\citep{im2flow2act} further introduces a data-efficient, fully autonomous flow-conditioned policy, leveraging task-agnostic datasets for one-shot real-world transfer.
Unlike these 2D methods, \modelname leverages 3D point-level flow, enhancing spatial localization and enables more accurate manipulation.
This approach builds upon prior work in 3D flow-based policies, which have shown promising results in articulated object manipulation~\citep{flowbot3d}, tool use~\citep{toolflownet,tooluseflow}, and general skill learning~\citep{generalflow}.
However, these methods typically rely on manually designed or heuristic policies during execution after estimating 3D flow.
In contrast, \modelname introduces an end-to-end manipulation policy, eliminating the need for heuristic post-processing.


\begin{figure*}[htbp]
\centering
\includegraphics[width=\linewidth]{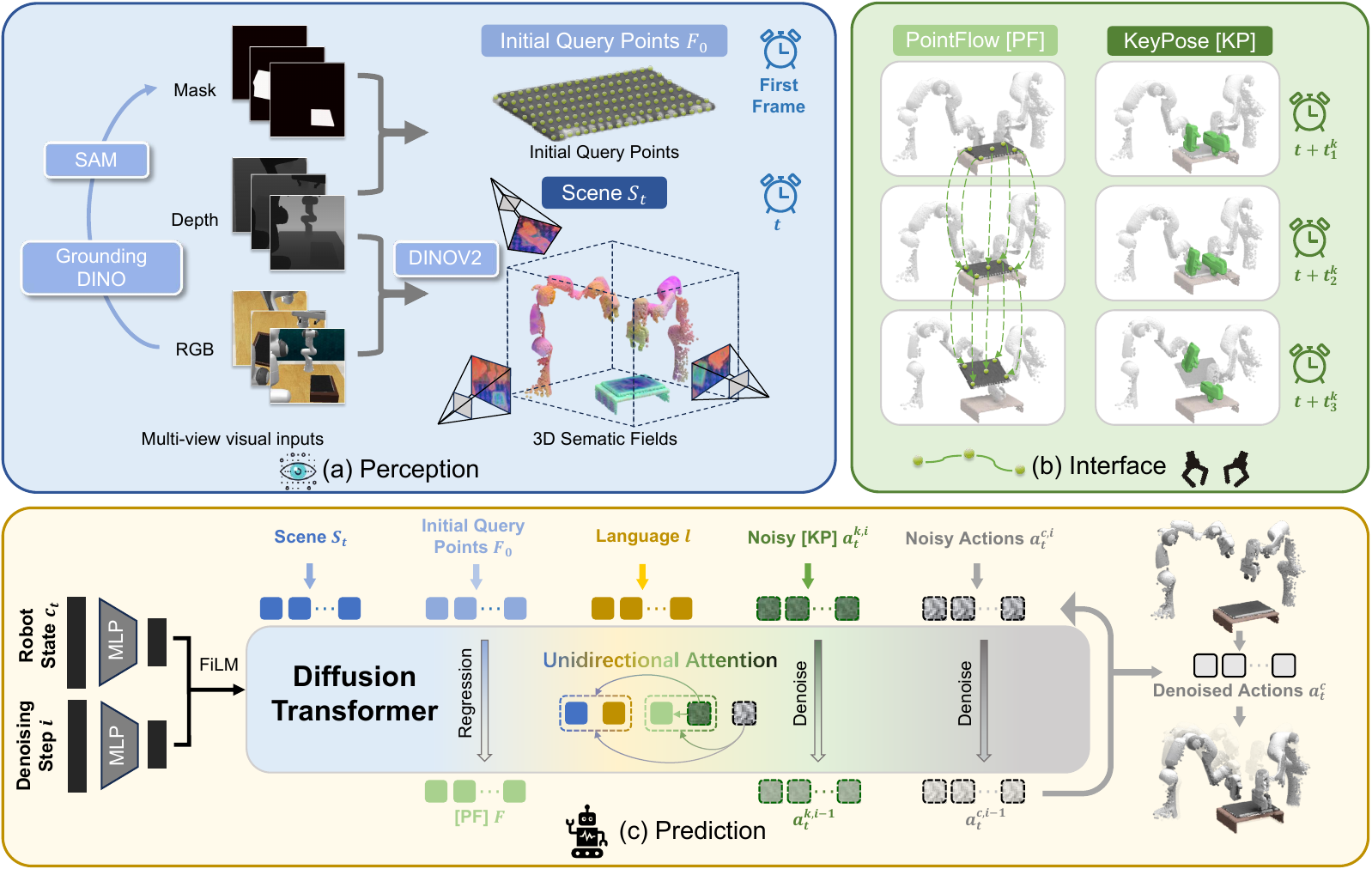}
\caption{
\textbf{Overview of \modelname.} 
(a) \textbf{Perception.} 
We first construct a 3D semantic neural field $S_t$ and sample initial query points $F_0$ for pointflow prediction.
(b) \textbf{Interface.} 
Next, we define two intermediate interfaces: target gripper poses $a_t^k$ and object pointflow $F$.
(c) \textbf{Prediction.}
Finally, a diffusion transformer incorporates robot proprio tokens $c_t$, scene tokens $S_t$, language tokens $l$, pointflow query tokens $F_0$ and  action tokens $a_t^k$ and $a_t^{c}$ with gaussian noise.
Using a carefully designed unidirectional attention, the model progressively denoises action predictions conditioned on the interfaces.
}
\label{fig:method}
\end{figure*}

\section{Method}
In this section, we describe \modelname in detail.
We begin with a brief problem formulation (Section \ref{problem formulation}). 
Next, we discuss the perception module (Section \ref{perception}) in \modelname, involving the construction of 3D semantic neural field and initial query points. 
Subsequently, we elaborate on the key interface designs—Pointflow and Keypose (Section \ref{Interface}), enhancing \modelname spatial localization capabilities.
Then, we illustrate the action prediction module (Section \ref{Prediction}), which is a diffusion-based transformer with unidirectional attention.
Finally, we provide a detailed implementation details during training and inference phases (Section \ref{implementation details}).


\subsection{Problem Formulation}\label{problem formulation}



%
At time step \( t \), \modelname takes inputs as the language instruction \( l \) and RGBD images $\mathcal{I}$ from \( K \) cameras, and outputs a sequence of $h^c$ continuous actions \( a_t^c = \{a_{t:t+h^c}\} \), where each action $a_t$ represents the target gripper poses and openness for both the left and right grippers.
Crucially, \modelname incorporates two intermediate \emph{\textbf{interfaces}} at \emph{\textbf{keyframe}} timesteps as additional conditions for action prediction. 
The \emph{\textbf{keyframes}} $t^k$ are defined as turning points in the trajectory where there are significant changes in the grippers' openness and the arms' joint states~\citep{qattention,perceiver}.
For the \emph{\textbf{interfaces}}, the first specifies the target gripper
poses at the subsequent \( h^k \) keyframes: \( a_t^k = \{a_{t^k_i}\}_{i=1}^{h^k} \). 
The second interface defines the positions of \( N_q \) spatial query points at the next \( h^k \) keyframes: \( F \in \mathbb{R}^{h^k \times N_q \times 3} \). 
At each keyframe timestep \( t^k_i \), the positions of the \( N_q \) points are denoted as \( F_{t^k_i} \in \mathbb{R}^{N \times 3} \), with initial positions at the first frame given by \( F_0 \in \mathbb{R}^{N \times 3} \).

\subsection{Perception} \label{perception}
\textbf{3D Scene Representation.} 
As is seen in Figure~\ref{fig:method}(a), we represent the scene using 3D semantic fields, focusing on both geometric and semantically meaningful regions. 
We begin by preprocessing the raw point clouds through cropping and downsampling.
For each sampled 3D point, we project it onto 2D RGB images from multiple camera viewpoints to extract pixel-wise semantic features using the DINOv2~\citep{dinov2} model.
We fuse these features through a weighted sum, where the weights are determined by the point's distance from the projected surface.
To mitigate the computational burden of numerous scene tokens in the transformer backbone, we downsample scene points while preserving their geometric and semantic information.
We use a PointNet++ dense encoder~\citep{pointnet} to obtain a compact scene representation \(S_t \in \mathbb{R}^{N_s \times (3+D)}\), where each of the \(N_s\) points encodes spatial coordinates and a D-dimensional fused semantic feature.
This compressed representation retains key geometric and semantic details while enhancing local point relationships through the set abstraction of PointNet++.

\textbf{Initial Query Points Sampling.} 
Instead of directly learning the pointflow distribution $p(F)$, we choose to approximate the conditional distribution $p(F|F_0)$, where $F_0$ represents the initial query points sampled from the manipulated object at the first frame.
This approach shifts the model’s focus from inferring global absolute coordinates to capturing overall object motion. 
Consequently, even when the object position is out of distribution, \modelname can estimate pointflow accurately and robustly, demonstrating improved generalization.

To get the initial query points $F_0$, we randomly sample $N_q$ query points from the object to be manipulated at the beginning of the task.
We use the Grounding DINO model~\citep{groundingdino} to obtain a bounding box from the language prompt and image, then input the bounding box into the SAM model~\citep{sam} to generate the object mask. 
We obtain the 3D coordinates $F_0 \in \mathbb{R}^{N_q \times 3}$ of the $N_q$ pixels sampled from the mask. 
In practice, we find that $N_q = 200$ points are sufficient for all tasks, both in simulation and the real world. 
It is worth noting that in each episode, we will only perform this operation once.

\subsection{Interface} \label{Interface}
%

%
\textbf{Target Gripper Poses.} 
We predict target gripper poses at keyframe timesteps as explicit action goals to better guide continuous action generation.
To supervise the target gripper poses, we first use a heuristic algorithm to identify the keyframe timesteps in the trajectory. 
Once a keyframe $t^k_i$ is established, its corresponding action label ${a}_{{t}^k_i}$ can be directly retrieved. 
If there are fewer keyframes remaining after the current timestep than  \( h^k \), we pad the sequence by repeating the action of the last keyframe.

\textbf{Object Pointflow.} 
A key challenge in obtaining ground truth labels for pointflow is the inevitable occlusion that occurs when the object moves or is manipulated.
We address this challenge by leveraging the object's 6D pose to track the real-time points' positions.
In simulation, we obtain the ground truth 6D pose label of rigid objects from the RLBench2~\citep{peract2} dataset, while in the real world, we estimate it using BundleSDF~\citep{bundlesdf} and Foundation Pose~\citep{foundationpose}.
Given the object's 6D pose at the first and each keyframe timestep,
we transform the query points from the first frame $F_0$ into the object’s coordinate frame and then back into world coordinates, yielding their keyframe positions \( F_{t^k_i} \in \mathbb{R}^{N \times 3} \), which serve as the ground truth for pointflow supervision.
Here real-time object 6D pose estimation is not required during inference.
Overall, object pointflow along with target gripper poses effectively model the interaction between the object and the robot.

\subsection{Prediction} \label{Prediction}
\textbf{Observation Encoder.}
\modelname processes four types of inputs: the 3D semantic neural field $S_t$, the language instruction $\ell$, the robot states $c_t$ and the initial positions of point queries $F_0$. 
The language is encoded using a CLIP text tokenizer~\citep{clip} and projected into a latent space via a three-layer MLP. 
Similarly, the robot states and point queries are projected into the latent space through a three-layer MLP.


\textbf{Diffusion Transformer.}
The backbone of the prediction module builds upon a diffusion transformer. 
At the time step t and denoising step $i$, let $a_t^{k,i}$ and $a_t^{c, i}$ be the keyframe action $a_t^{k}$ and continuous action $a_t^{c}$ with noise.
The transformer incoporates the scene tokens $S_t$, language tokens $l$, query points tokens $F_0$ and noised action tokens $a_t^{k, i}$ and $a_t^{c,i}$.
Outputs are supervised by gaussian noise $\epsilon_k^i$, $\epsilon_c^i$ via DDPM~\citep{ddpm} training and ground truth pointflow $F$ via direct regression.

Notably, we design a unidirectional attention that leverages the interfaces to bridge the gap between input and output modalities.
As shown in Figure~\ref{fig:method}(c), all pointflow and action tokens attend to the scene and language tokens, integrating spatial and semantic knowledge. 
Moreover, the noised keyframe action token $a_t^{k,i}$ attends to the pointflow token, aiming to extract additional object-level features.
The final continuous action token $a_t^{c, i}$  attends to all the previous tokens, not only distilling regular scene-level features but also fully utilizing the local and detailed features contained in the interfaces.


%
We apply relative attention, as introduced in previous work~\citep{3d-diffuser-actor}, between point flow tokens $F$, keyframe action tokens $a_t^{k,i}$, continuous action tokens $a_t^{c,i}$, and scene point cloud tokens $S_t$, enabling the encoding of relative 3D positional information in the attention layers. This relative attention relies on the relative 3D positions of features and is implemented using rotary positional embeddings~\cite{rotary}.
For language instructions $l$, we use regular cross and self-attention. 
The robot's proprioception $c_t$ and denoising timesteps $i$ affect the attention through Feature-wise Linear Modulation(FiLM)~\citep{film}.


\begin{figure*}[htbp]
\centering
\includegraphics[width=\linewidth]{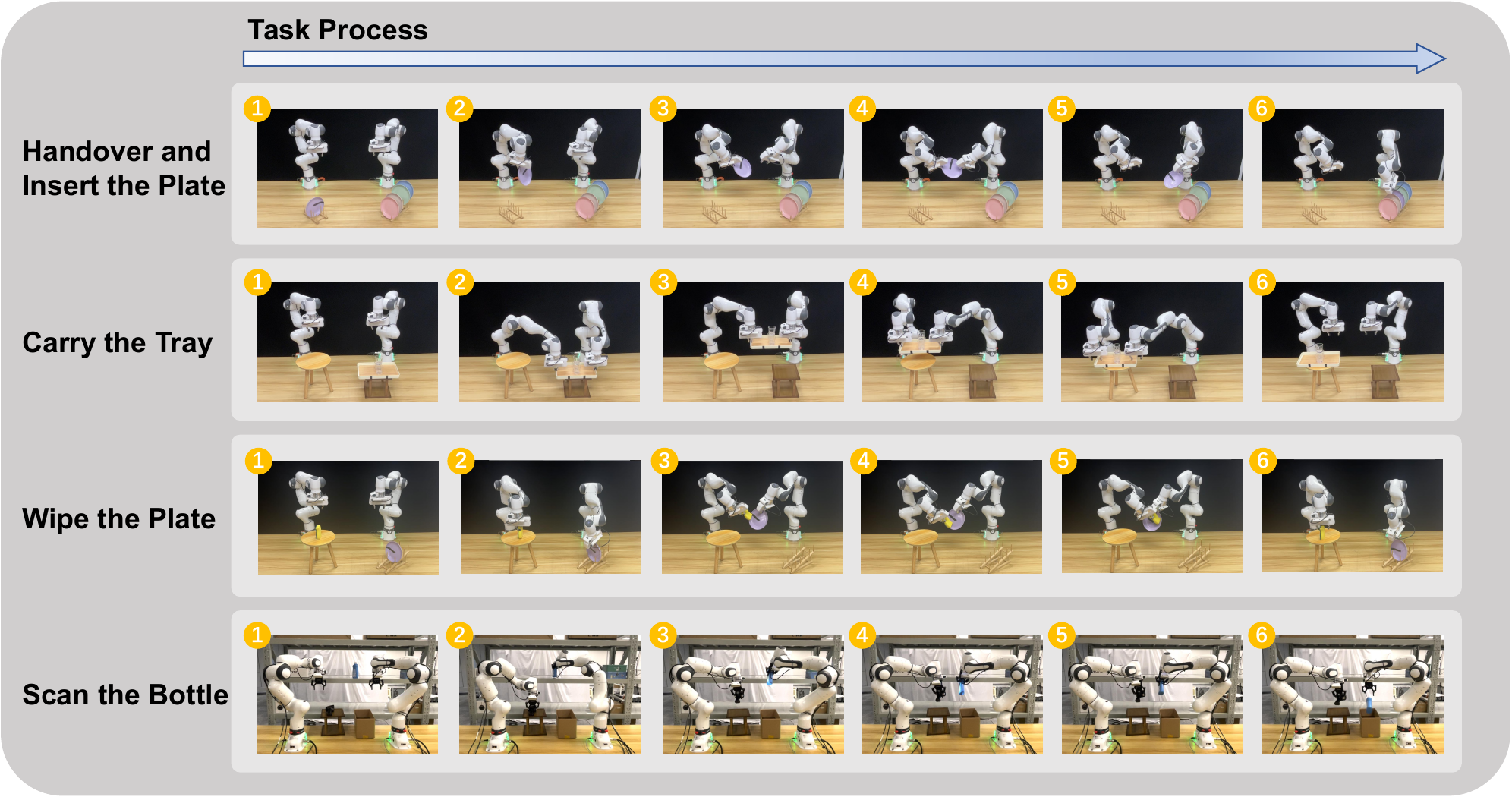}
\caption{Task process visualizations in four real-world tasks.}
\label{Task process Figure}
\end{figure*}

\subsection{Implementation Details}\label{implementation details}
\textbf{Training.}
For the \modelname network \( \epsilon_{\theta} \), the continuous action loss $\mathcal{L}_{c}$, the keyframe action loss $\mathcal{L}_{k}$, and the  pointflow prediction loss $\mathcal{L}_{F}$ are computed via $L_1$ loss.
\[\mathcal{L}_c=||\epsilon_{\theta}(S_t,F_0,\ell,c_t,a_t^{c,i},i)-\epsilon_c^i||\]
\[\mathcal{L}_k=||\epsilon_{\theta}(S_t,F_0,\ell,c_t,a_t^{k,i},i)-\epsilon_k^i||\]
\[\mathcal{L}_{F}=||\epsilon_{\theta}(S_t,F_0,\ell,c_t,i)-F||\]
%
The overall training loss is:
\[\mathcal{L}_{\theta}=w_1 \mathcal{L}_{c}+ w_2\mathcal{L}_{k} + w_3\mathcal{L}_{F}\]
where $w_1$, $w_2$, and $w_3$ are hyperparameters set to 0.05, 0.05, and 1, respectively.
We use DDPM with 1000 training timesteps for noise scheduling in all experiments. 
We train 500 epochs for tasks from RLBench2 benchmark and 5000 epochs in real-world tasks, with a batch size of 128 and learning rate of 1e-4 with a cosine decay learning rate scheduler. 
For tasks involving 100 episodes, each with 150-250 timesteps, we train the model using eight A100 GPUs for approximately 20 hours, and use the checkpoint with the lowest average validation loss for evaluation.

\textbf{Inference.}
We begin by sampling 200 points on objects as query points at the first timestep. 
Then at each timestep $t$, we draw random initial continuous action samples \( a_t^{c} \) and keyframe action samples \( a_t^{k} \) from a Gaussian distribution and denoise 1000 steps with DDPM in simulation tasks and 20 steps with DDIM\cite{ddim} for real-world tasks. 
In practice, we predict 50 continuous actions and 4 keyframe actions and pointflow.

\section{Real-world Experiments}

We carefully design four real-world tasks with high localization demands and motion constraints to evaluate \modelname’s capabilities in:
(1) Effectiveness in long-horizon tasks.
(2) Robustness under high-intensity disturbances in objects and environments.

\subsection{Real-world Experiments Setup}

\begin{figure}[htbp]
    \centering
    \includegraphics[width=\linewidth]{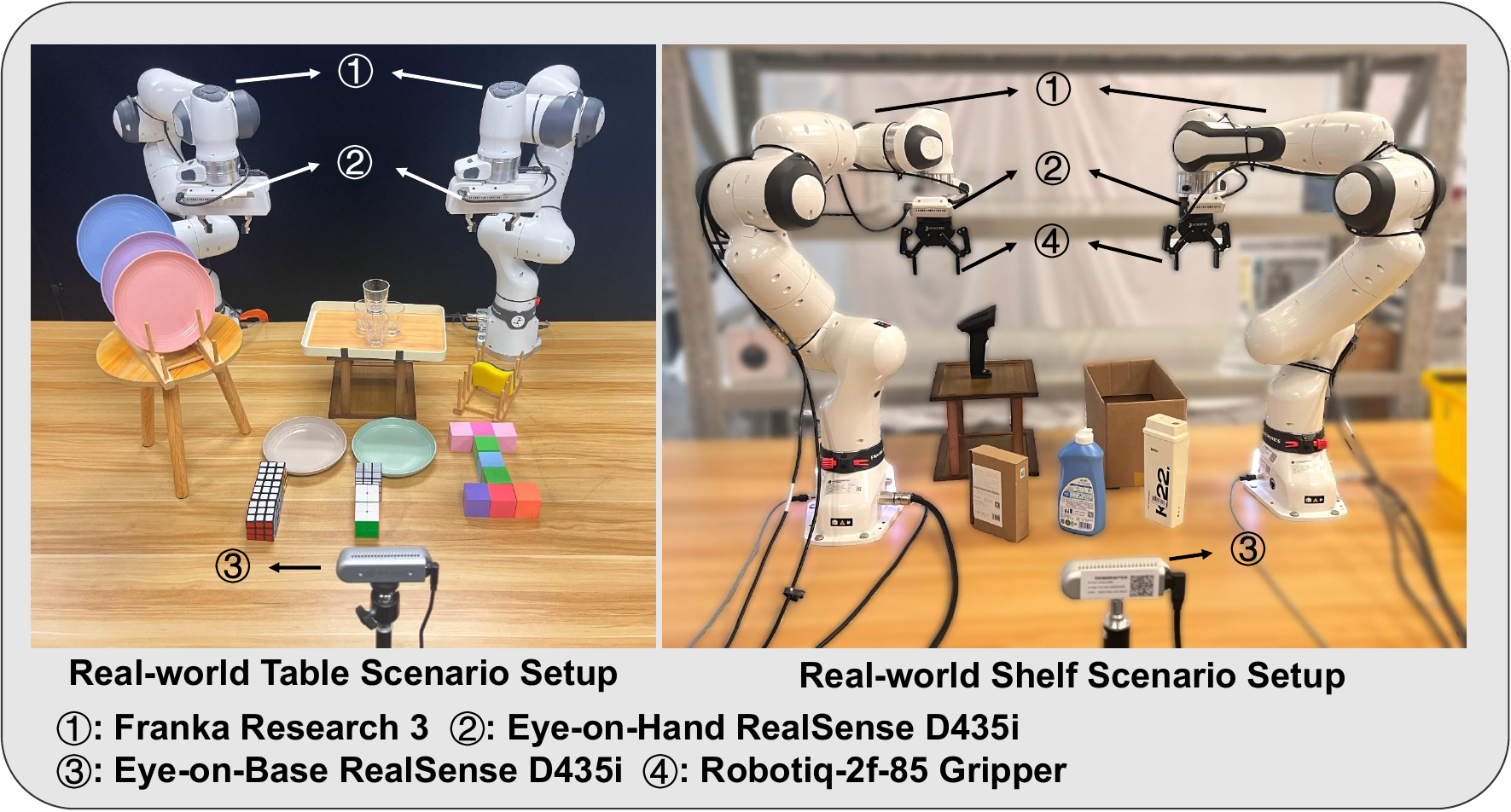}
    \caption{Two real-world setups.}
    \label{fig:real-settings}
\end{figure}

\textbf{Benchmark.}
We evaluate our method on two Franka Research 3 robots across four tasks in two distinct scenarios (Figure~\ref{fig:real-settings}).
Each scene is equipped with two Eye-on-Hand and one Eye-on-Base RealSense D435i cameras.
In the first desktop environment, we test three long-horizon tasks requiring high localization accuracy and curved motion execution.
In the second shelf scenario, we employ Robotiq-2f-85 grippers as end-effectors for the final task.
Below, we briefly outline the process for each of the four tasks (Figure~\ref{Task process Figure}), with further details provided in the Appendix.

\begin{table*}[htbp]
\centering
\caption{\textbf{Real-world main results.}
We evaluate all the methods with 10 (settings) × 3 (repeated trials)
rollouts per task. Our method achieves better performances among all tasks than baselines.
The best results are bolded.
}
\label{real-main}
\scalebox{0.93}{




\begin{tabular}{ccccccc}
\toprule

\multirow{2}{*}{Method} & 
\multicolumn{6}{c}{SR (\%) $\uparrow$ /Loc-SR (\%) $\uparrow$ / Normalized Score $\uparrow$} \\
\cline{2-7}
& Avg. Success $\uparrow$ & Carry the Tray  & Handover and Insert the Plate &  Wipe the Plate  & Scan the Bottle \\
\midrule \midrule 
ACT 
& 15.0 / 50.0 / 4.7
& 40.0 / 40.0 / 5.5
& 10.0 / 70.0 / 5.0
& 0.00 / 40.0 / 3.8 
& 10.0 / 50.0 / 4.3    
& \\
DP3 
& 35.0 / 57.5 / 5.5
& 50.0 / 50.0 / 6.0
& 20.0 / 80.0 / 4.7
& 40.0 / 60.0 / 6.5
& 30.0 / 40.0 / 4.8
&  \\
3D Diffuser Actor 
& 5.00 / 65.0 / 4.6
& 0.00 / 70.0 / 4.5
& 0.00 / 100 / 6.0
& 0.00 / 50.0 / 3.5
& 20.0 / 40.0 / 4.8 
& \\
\baseline{Ours}
& \baseline{\textbf{62.5} / \textbf{92.5} / \textbf{8.2}}
& \baseline{\textbf{50.0} / \textbf{100} / \textbf{7.8}}
& \baseline{\textbf{40.0} / \textbf{100} / \textbf{7.7}} 
& \baseline{\textbf{70.0} / \textbf{80.0} / \textbf{8.3}}
& \baseline{\textbf{90.0} / \textbf{90.0} / \textbf{9.3}}
& \\
\bottomrule
\end{tabular}}
\vspace{-5mm}
\end{table*}

\begin{enumerate}
    \item \textbf{Carry the Tray.}
    Both arms must collaboratively lift a tray stacked with cups or other objects from a lower platform and steadily transfer it to a small table.
    This task evaluates the ability to precisely locate the tray’s centerline and maintain stable, coordinated movements throughout the process.
    \item \textbf{Handover and Insert the Plate.} %
    The right arm picks up the plate and hands it over to the left arm, which then inserts it into an available slot in the rack.
    This task tests temporal coordination during handover and precise spatial perception.
    \item \textbf{Wipe the Plate.}
    Each arm picks up the sponge and the plate, respectively, and uses the sponge to wipe the plate.
    After wiping, both objects are returned to their original positions.
    This task evaluates the ability to perform curved motions in interactive tasks.
    \item \textbf{Scan the Bottle.} 
    The right arm retrieves the bottle from the shelf, while the left arm picks up the scanner from the table to scan the bottle’s barcode.
    After scanning, the bottle is placed into a box.
    This task assesses 6-DoF picking in a more constrained spatial environment and the coordination between both arms.
\end{enumerate}

\textbf{Expert demonstrations.}
We constructed two isomorphic teleoperation devices GELLO~\citep{wu2024gello} for the Franka Research 3 to collect expert demonstrations.
%
%
We collected 50 demonstrations for the task ``Carry the Tray" and 20 for other tasks. The limited number of training samples is intended to evaluate whether the policies achieve excellent spatial localization and perception of objects with minimal data.

\textbf{Baselines.}
We implement an Action Chunking Transformer (ACT)~\citep{act} that predicts target joint positions from 2D RGB inputs. 
We also adapt DP3~\citep{dp3} into a bimanual framework, which is a 3D point-cloud-based continuous control policy. 
Additionally, we reproduce the 3D Diffuser Actor~\citep{3d-diffuser-actor}, a keyframe-based diffusion policy utilizing 3D semantic fields.

\textbf{Metrics.}
Each method is evaluated across 10 settings, with 3 trials per setting. 
Due to the long-horizon and complex nature of the tasks, we have established three key metrics: Success Rate (SR), Localization Success Rate (Loc-SR), and Normalized Score.
The Success Rate (SR) is only assigned a value of 100\% upon the successful completion of the entire task.
Loc-SR will be recorded 100\% if the robots perform well to find the object contact positions.
Additionally, we divide each tasks into 3 or 4 intermediate stages, and the Score is progressively accumulated through the completion of individual intermediate stages.
Given that the number of steps varies across these tasks, the score will be normalized to a scale of 10. 
The scoring criteria and other details are available in the Appendix.



\subsection{Real-world Main Results}\label{Real-world Main Results}

As shown in Table~\ref{real-main}, \modelname outperforms all baselines across tasks.
Compared to state-of-the-art methods, it achieves a 27.5\% increase in average success rate (SR), a 27.5\% improvement in localization success rate (Loc-SR), and a 2.7 point gain in Normalized Score.
Compared to ACT, which relies on RGB inputs, \modelname demonstrates superior Loc-SR and overall scores.
As a single-frame observation algorithm, ACT struggles with tasks requiring repetitive trajectories (e.g.,``Wipe the Plate"), whereas \modelname integrates multistep proprioception, effectively leveraging historical information.
Compared to DP3, a point-cloud-based method prone to overfitting seen trajectories and susceptible to noise~\citep{rise}, \modelname exhibits stronger localization ability and robustness to noised pointcloud.
While 3D Diffuser Actor, a keyframe-based policy using semantic neural fields and heuristic action execution, performs well in Loc-SR, it fails to account for collision constraints, leading to a lower SR.
In contrast, \modelname not only inherits the perception advantages of keyframe prediction but also effectively respects path constraints inherent in training demonstrations.

\subsection{Real-world Generalization Tests}
\begin{figure}[htbp]
    \centering
    \includegraphics[width=\linewidth]{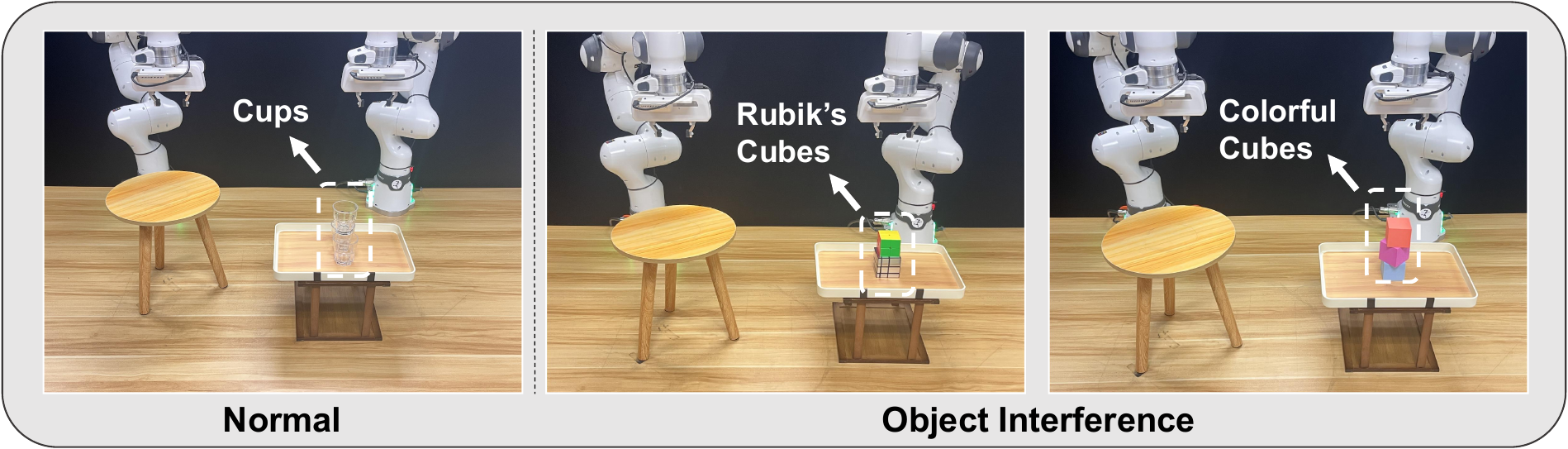}
    \caption{Visualization under object interference in Carry the Tray.}
    \label{fig:general-carry}
\vspace{-3mm}
\end{figure}

\begin{figure}[htbp]
    \centering
    \includegraphics[width=\linewidth]{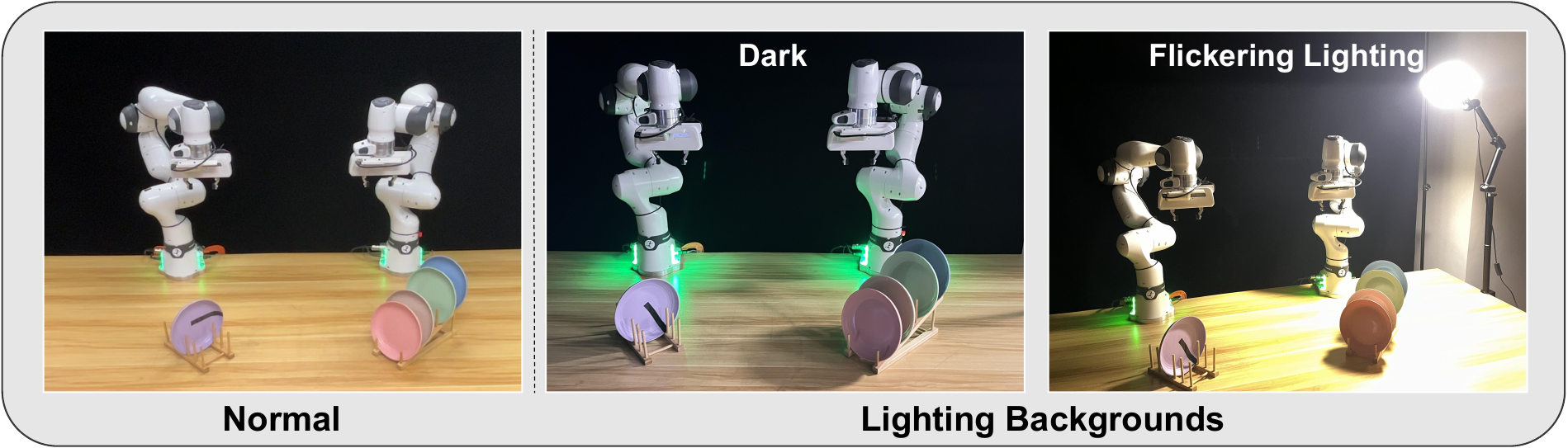}
    \caption{Visualization under different lighting backgrounds in Handover and Insert the Plate.}
    \label{fig:general-handover}
\vspace{-3mm}
\end{figure}

\begin{figure}[htbp]
    \centering
    \includegraphics[width=\linewidth]{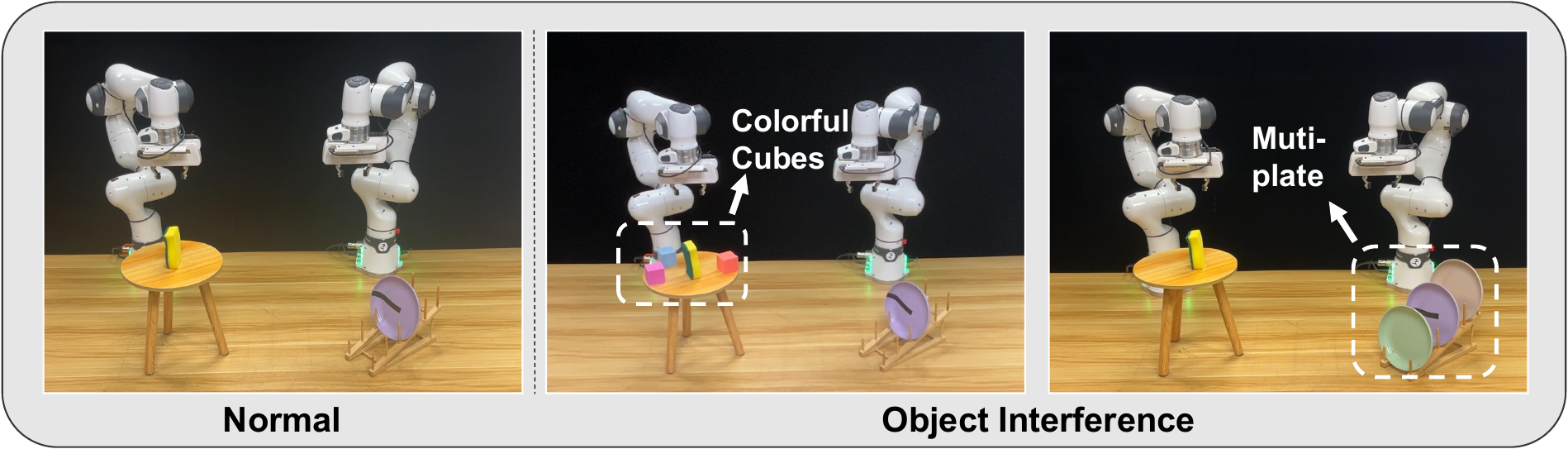}
    \caption{Visualization under object interference in Wipe the Plate.}
    \label{fig:general-wipe}
\vspace{-3mm}
\end{figure}

\begin{figure}[htbp]
    \centering
    \includegraphics[width=\linewidth]{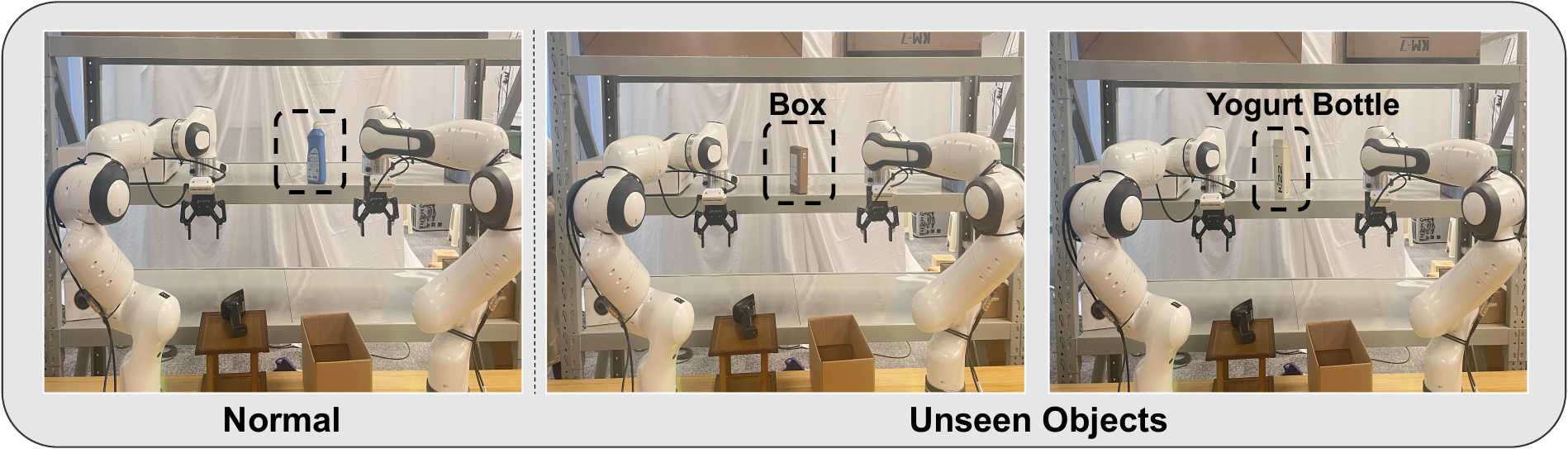}
    \caption{Visualization under unseen objects in Scan the Bottle.}
    \label{fig:general-scan}
\vspace{-3mm}
\end{figure}

As shown in Table~\ref{gen-carry}-\ref{gen-scan} and Figure~\ref{fig:general-carry}-\ref{fig:general-scan}, We introduce three types of generalizations to evaluate the generalizability and robustness: unseen objects to manipulated, different lighting backgrounds and interference from other objects. 
In each task, we select the setting where each method performed the best for generalization testing, and then we record the success rate, localization success rate, and normalized score. 
Each different generalization scenario has 3 trials.
Details results are available in the Appendix.

\begin{table}[htbp]
\centering
\caption{Evaluation under object interference in Carry the Tray.}
\label{gen-carry}
\resizebox{\linewidth}{!}{\begin{tabular}{cccc}
\toprule
\multirow{2}{*}{Method} & \multicolumn{3}{c}{Success / Localization / Normalized Score} \\ \cline{2-4}
 & Normal Setting 
 & Rubik’s Cubes 
 & Colorful Cubes 
 \\ 
 \midrule \midrule 
ACT 
& \greencheckmark / \greencheckmark / 10
& \greencheckmark / \greencheckmark / 10
& \greencheckmark / \greencheckmark / 10
\\
DP3 
& \greencheckmark / \greencheckmark / 10
& \redxmark / \redxmark / 2.5
& \redxmark / \greencheckmark  / 5.0
\\
3D Diffuser Actor
& \redxmark / \greencheckmark / 7.5
& \redxmark / \greencheckmark / 5.0
& \redxmark / \redxmark / 2.5
\\
\baseline{Ours}
& \baseline{\greencheckmark / \greencheckmark / 10}
& \baseline{\greencheckmark / \greencheckmark / 10}
& \baseline{\greencheckmark / \greencheckmark / 10}
\\
\bottomrule
\end{tabular}
}
\vspace{-3mm}
\end{table}
In \textbf{Carry the Tray}, we replace the cups with Robik's Cubes and colorful cubes to test robustness against \textbf{object interference}.
These objects differ in size and color, which not only affect RGB inputs but also alter depth, influencing 3D-based policies.
Thanks to the two interfaces, particularly the pointflow, \modelname focuses on key object-related regions, such as the tray, without being distracted by other objects that could affect localization.
Interestingly, two 3D-based baselines perform poorly, while the 2D-based ACT achieves better generalization.
This suggests that although objects in this task have minimal impact at the pixel level in 2D images, they have a greater effect on the 3D scene representation.

\begin{table}[htbp]
\centering
\caption{Evaluation under different lighting backgrounds in Handover and Insert the Plate.}
\label{gen-handover}
\resizebox{\linewidth}{!}{\begin{tabular}{cccc}
\toprule
\multirow{3}{*}{Method} & \multicolumn{3}{c}{Success / Localization / Normalized Score} \\ \cline{2-4}
 & Normal Setting 
 & \makecell{Dark \\ Environment}
 & \makecell{Flickering Lighting \\ Environment}
 \\ 
 \midrule \midrule 
ACT 
& \greencheckmark / \greencheckmark / 10
& \redxmark / \redxmark / 0.0
& \redxmark / \redxmark / 0.0
\\
DP3 
& \greencheckmark / \greencheckmark / 10
& \redxmark / \greencheckmark / 6.7
& \greencheckmark / \greencheckmark  / 10
\\
3D Diffuser Actor
& \redxmark / \greencheckmark / 6.7
& \redxmark / \redxmark / 0.0
& \redxmark / \redxmark / 0.0
\\
\baseline{Ours}
& \baseline{\greencheckmark / \greencheckmark / 10}
& \baseline{\greencheckmark / \greencheckmark / 10}
& \baseline{\greencheckmark / \greencheckmark / 10}
\\
\bottomrule
\end{tabular}
}
\vspace{-3mm}
\end{table}

In \textbf{Handover and Insert the Plate}, we evaluate \modelname’s adaptability to varying \textbf{lighting conditions}.
In dark and flickering lighting environments, RGB-based information is severely affected.
However, despite using color information as inputs, \modelname’s performance remains stable.
This robustness probably stems from its reliance on pointflow and keypose, both derived mostly from the integration of semantic and positional features rather than pure color information.
Meanwhile, DP3 maintains some success, likely due to its use of colorless point clouds.

\begin{table}[htbp]
\centering
\caption{Evaluation under object interference in Wipe the Plate.}
\label{gen-wipe}
\resizebox{\linewidth}{!}{\begin{tabular}{cccc}
\toprule
\multirow{2}{*}{Method} & \multicolumn{3}{c}{Success / Localization / Normalized Score} \\ \cline{2-4}
 & Normal Setting 
 & Colorful Cubes 
 & Multi-plate
 \\ 
 \midrule \midrule 
ACT 
& \redxmark / \greencheckmark / 7.5
& \redxmark / \redxmark / 2.5
& \redxmark / \redxmark / 2.5
\\
DP3 
& \greencheckmark / \greencheckmark / 10
& \redxmark / \redxmark / 0.0
& \redxmark / \redxmark  / 0.0
\\
3D Diffuser Actor
& \redxmark / \greencheckmark / 5.0
& \redxmark / \greencheckmark / 5.0
& \redxmark / \greencheckmark / 5.0
\\
\baseline{Ours}
& \baseline{\greencheckmark / \greencheckmark / 10}
& \baseline{\greencheckmark / \greencheckmark / 10}
& \baseline{\redxmark / \greencheckmark / 7.5}
\\
\bottomrule
\end{tabular}
}
\vspace{-3mm}
\end{table}
In \textbf{Wipe the Plate}, we introduce \textbf{object interference} separately for the objects manipulated by each arm, leading to partial occlusions and various visual distractions.
Despite this, only \modelname maintains robust localization.


In \textbf{Scan the Bottle}, we evaluate our method’s generalization to \textbf{unseen objects}.
Notably, \modelname enables zero-shot manipulation of new objects by adjusting GroundingDino’s~\citep{groundingdino} prompt to obtain novel initial query points $F_0$ for pointflow prediction.
%
%
This is driven by \modelname’s learned conditional distribution $p(F|F_0)$, which prioritizes object motion changes over absolute global coordinates $p(F)$.
As a result, even when encountering novel rigid objects, \modelname effectively estimates a rough object's relative transformations, enhancing task completion.

\begin{table}[htbp]
\centering
\caption{Evaluation under unseen objects in Scan the Bottle.}
\label{gen-scan}
\resizebox{0.9\linewidth}{!}{\begin{tabular}{cccc}
\toprule
\multirow{2}{*}{Method} & \multicolumn{3}{c}{Success / Localization / Normalized Score} \\ \cline{2-4}
 & Normal Setting 
 & Box
 & Yogurt bottle
 \\ 
 \midrule \midrule 
ACT 
& \greencheckmark / \greencheckmark / 10
& \redxmark / \redxmark / 2.5
& \redxmark / \redxmark / 2.5
\\
DP3 
& \greencheckmark / \greencheckmark / 10
& \redxmark / \redxmark / 2.5
& \redxmark / \redxmark  / 2.5
\\
3D Diffuser Actor
& \greencheckmark / \greencheckmark / 10
& \redxmark / \redxmark / 2.5
& \redxmark / \redxmark / 2.5
\\
\baseline{Ours}
& \baseline{\greencheckmark / \greencheckmark / 10}
& \baseline{\redxmark / \greencheckmark / 7.5}
& \baseline{\redxmark / \greencheckmark / 7.5}
\\
\bottomrule
\end{tabular}
}
\vspace{-3mm}
\end{table}


\section{Simulation Experiments}
We evaluate our approach on seven tasks using the bimanual simulation benchmark RLBench2~\citep{peract2}, aiming to address the following questions: (1) How well does \modelname perform on complex bimanual tasks? (2) Are the proposed target gripper poses and object pointflow interfaces effective? (3) How do these interfaces learn scene features to enhance guidance for continuous action prediction?

\subsection{Simulation Experiments Setup}

\begin{table*}[htbp]
\centering
\caption{\textbf{Quantitative results on RLBench2.} For each task, we present the average performance of three
checkpoints averaged over 100 rollouts. The metric “Avg. Success” measures the average success
rate across seven tasks. \modelname outperforms baselines with higher Avg. Success and better results on
most tasks. The best results are bolded.}
\label{sim-main}
\scalebox{0.9}{\begin{tabular}{cccccccccc}
\toprule
Method 
& Avg. Success $\uparrow$ 
& Box 
& Ball 
& Drawer  
& Laptop  
& Dustpan 
& Tray   
& Handover Easy  \\
\midrule \midrule 
ACT 
& 15.4
& 67.0 \textcolor{gray!100}{\tiny$\pm 7.0$}
& 38.3 \textcolor{gray!100}{\tiny$\pm 10.0$}
& 1.7 \textcolor{gray!100}{\tiny$\pm 2.1$}
& 0.0\textcolor{gray!100}{\tiny$\pm 0.0$} 
& 0.0 \textcolor{gray!100}{\tiny$\pm 0.0$}
& 1.3 \textcolor{gray!100}{\tiny$\pm 1.5$}
& 0.0 \textcolor{gray!100}{\tiny$\pm 0.0$}\\
DP3 
& 26.0
& 39.3 \textcolor{gray!100}{\tiny$\pm 3.1$}
& 27.0 \textcolor{gray!100}{\tiny$\pm 6.6$}
& 0.0 \textcolor{gray!100}{\tiny$\pm 0.0$}
& 6.0 \textcolor{gray!100}{\tiny$\pm 2.6$}
& \textbf{98.7} \textcolor{gray!100}{\tiny$\pm 0.6$}
&6.3 \textcolor{gray!100}{\tiny$\pm 0.6$} 
& 4.7 \textcolor{gray!100}{\tiny$\pm 1.5$} \\
3D Diffuser Actor 
& 64.7
& 54.7 \textcolor{gray!100}{\tiny$\pm 3.8$} 
& 87.3 \textcolor{gray!100}{\tiny$\pm 1.9$} 
& 52.7 \textcolor{gray!100}{\tiny$\pm 14.1$}
& 40.7 \textcolor{gray!100}{\tiny$\pm 6.1$}
& 96.7 \textcolor{gray!100}{\tiny$\pm 2.6$} 
& 76.0 \textcolor{gray!100}{\tiny$\pm 4.3$} 
& 44.7 \textcolor{gray!100}{\tiny$\pm 3.3$}\\
PerAct$^2$ 
& 40.0
& 62.0  \textcolor{gray!100}{\tiny$\pm 26.2 $} 
& 50.0 \textcolor{gray!100}{\tiny$\pm 8.7$} 
& 49.7 \textcolor{gray!100}{\tiny$\pm 16.6$}
& 36.7 \textcolor{gray!100}{\tiny $\pm 5.7$}
& 2.0 \textcolor{gray!100}{\tiny$\pm 3.5$} 
& 60.0 \textcolor{gray!100}{\tiny$\pm 6.2$} 
& 19.7\textcolor{gray!100}{\tiny$\pm 6.0$}\\
\baseline{Ours} 
& \baseline{\textbf{80.8}}
& \baseline{\textbf{96.7} \textcolor{gray!100}{\tiny$\pm 1.5$} }
& \baseline{\textbf{89.3} \textcolor{gray!100}{\tiny$\pm 1.5$} }
& \baseline{\textbf{79.7} \textcolor{gray!100}{\tiny$\pm 3.8$}}
& \baseline{\textbf{46.3} \textcolor{gray!100}{\tiny $\pm 1.2$}}
& \baseline{\textbf{98.7} \textcolor{gray!100}{\tiny$\pm 1.5$} }
& \baseline{\textbf{92.0} \textcolor{gray!100}{\tiny$\pm 1.0$} }
& \baseline{\textbf{62.7}\textcolor{gray!100}{\tiny$\pm 2.5$}}\\
\bottomrule
\end{tabular}}
\end{table*}
\begin{table*}[htbp]
\centering
\caption{\textbf{Ablation studies.}
For all ablated models, we report best performance, while for \modelname, we additionally present the average performance across three checkpoints.
Overall, integrating both keypose and pointflow achieves the highest performance.}
\label{sim-ablation}
\scalebox{1}{\begin{tabular}{ccccccccc}
\toprule
Method & Avg. Success $\uparrow$ & Box & Ball &  Drawer  & Laptop  &  Dustpan & Tray   & Handover Easy  \\
\midrule \midrule 
Continuous & 47.6 & 84 & 24 & 41 &  0& 98& 82& 4\\
Keyframe & 49.0 & 84 & \textbf{94} & 36 & 0&92& 12& 25 \\
Continuous on Keypose & 53.6 & 71& 81& 29& 1& 99& 86& 8 \\
Continuous on Pointflow & 74.3 & 92 &77 &\textbf{84} & 29& 99&89 & 50\\

\midrule \midrule
\baseline{Ours (Best Ckpt)} 
& \baseline{\textbf{82.6}}
& \baseline{\textbf{98}}
& \baseline{91}
& \baseline{\textbf{84}}
& \baseline{\textbf{47}}
& \baseline{\textbf{100}}
& \baseline{\textbf{93}}
& \baseline{\textbf{65}} \\
\baseline{Ours (Averaged)} 
& \baseline{80.8}
& \baseline{96.7}
& \baseline{89.3}
& \baseline{79.7}
& \baseline{46.3}
& \baseline{98.7}
& \baseline{92.0}
& \baseline{62.7} \\
\bottomrule
\end{tabular}}
\end{table*}

\textbf{Benchmark.}
RLBench2~\citep{peract2} is a bimanual manipulation benchmark built on CoppeliaSim, encompassing tasks with different levels of coupling, coordination, language instructions, and manipulation skills. 
We select seven representative and challenging tasks and regenerate the training data. 
The official dataset suffers from significant misalignment between training and evaluation, such as robot shadows present in the training set but absent during evaluation.
Additionally, it lacks meta information for acquiring object pointflow.

\textbf{Baselines.}
We use the same three baselines as in the real-world experiments: DP3, ACT, and 3D Diffuser Actor. 
Additionally, we include PerAct$^2$~\citep{peract2}, a state-of-the-art method previously reported on RLBench2. 
It employs a Perceiver architecture to voxelize 3D spaces and predict keyframe actions.

%

\textbf{Metrics.}
Each method is evaluated across 100 rollouts per task with varying initial states for each tasks.
We report both per-task and average success rates, with all performances computed from three different checkpoints.

\subsection{Simulation Main Results}

As shown in Table~\ref{sim-main}, \modelname achieves an average success rate of 80.8\%, significantly outperforming the baselines. 
Compared to ACT (2D-based algorithm), \modelname leverages 3D semantic neural fields, and provides superior spatial perception.
While both methods utilize 3D point clouds, \modelname outperforms DP3 by 54\% in success rate, demonstrating that \modelname's spatial information processing is more useful than DP3's approach of encoding the entire scene into a single token.

Moreover, ACT and DP3, as continuous-action-based policies, underperforms keyframe-based policies like 3D Diffuser Actor and PerAct$^2$ by a margin of at least 10\%.
This disparity arises from keyframe-based methods offering more effective perception capabilities rather than overfitting to trajectories.
By using keypose as an interface, \modelname integrates the spatial awareness of keyframe methods with semantic neural fields, achieving substantial improvements over continuous-action-based approaches.


%
Beyond that, keyframe-based policies like 3D Diffuser Actor and PerAct$^2$ fall behind our model in tasks with movement constraints, such as the lift tray task (which requires both hands to remain level) and the push box task (which involves continuously pushing).
This is because our policy provides greater flexibility in managing such constraints, enabled by \modelname’s supervision of continuous actions between keyframes.
Additionally, in tasks demanding high spatial precision, such as object-picking tasks like Drawer and Handover Easy, our model significantly outperforms baselines. 
This advantage stems primarily from the use of object pointflow as an interface, which enhances localization accuracy.

\begin{figure*}[htbp]
    \centering
    \includegraphics[width=0.9\linewidth]{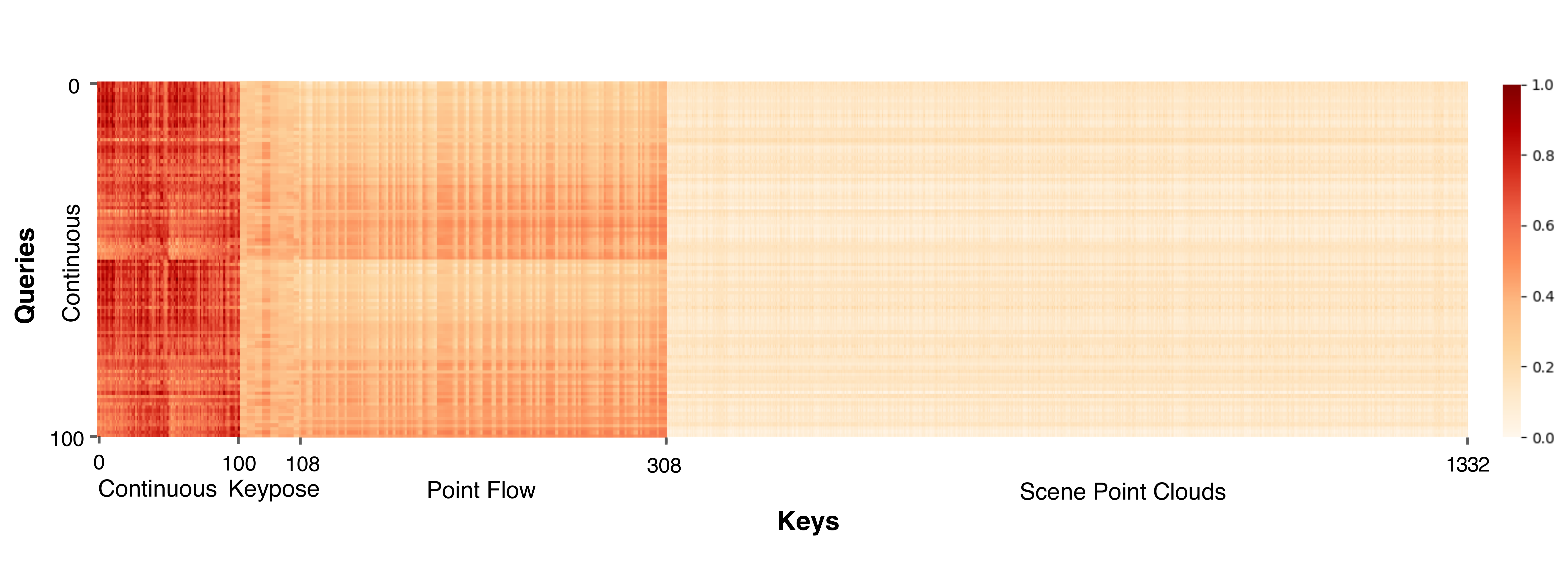}
    \vspace{-3mm}
    \caption{\textbf{Heatmap of the attention weights of continuous action tokens.} The y-axis represents continuous action tokens as queries. The x-axis sequentially displays continuous action, keypose, pointflow, and scene tokens as keys.}
    \label{fig:general-heatmap}
    \vspace{-3mm}
\end{figure*}

\subsection{Ablation Study}
As shown in Table~\ref{sim-ablation}, we analyze the contributions of target gripper poses and object pointflow interfaces on RLBench2 by comparing different ablated models at their best performance.

We begin by evaluating the vanilla keyframe and continuous baselines, which predict only keypose or continuous actions, respectively.
In tasks requiring precise positioning, such as Handover Easy and Lift Ball, the keyframe-based policy demonstrates superior localization.
However, in tasks involving horizontal lifting (Lift Tray) or curved motion trajectories (Sweep to Dustpan), the continuous-based approach significantly outperforms the keyframe method.
These results indicate that relying solely on either keyframe or continuous actions is insufficient for general manipulation tasks.

Next, we modify the continuous-based policy by incorporating target gripper keypose and object pointflow predictions (line 3 and 4 in Table~\ref{sim-ablation}). 
Conditioning on separate interfaces improves performance, likely due to the local spatial features they provide.
Moreover, combining both interfaces yields further gains, highlighting the synergy between keypose and pointflow in enhancing performance on downstream tasks.
%


\subsection{Visualization Analysis}
%
In this section, we intuitively analyze how the proposed interfaces enhance localization accuracy and generalization to distractors and task-irrelevant backgrounds.

\begin{figure}[htbp]
    \centering
    \includegraphics[width=\linewidth]{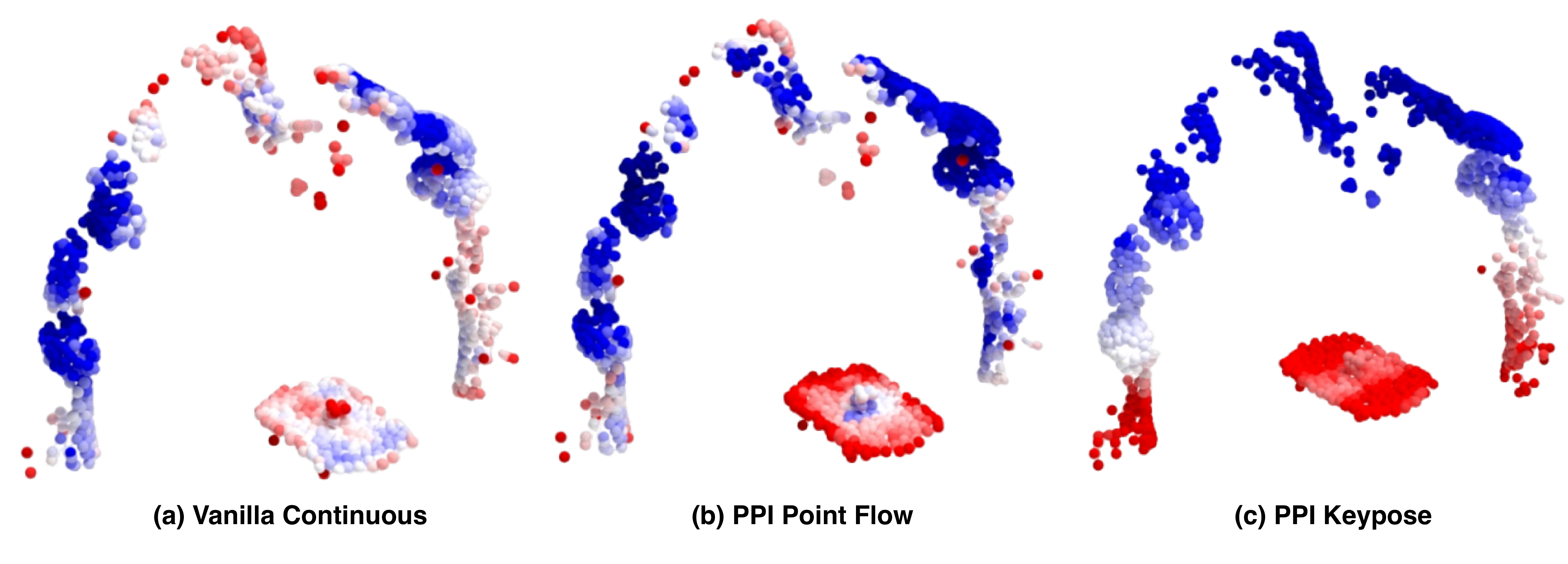}
    \caption{\textbf{Visualization of the attention weights of interface tokens.} We use ``Lift Tray'' task on RLBench2 as an example. The \textcolor{red}{\textbf{red}} area corresponds to larger attention weight, while the \textcolor{blue}{\textbf{blue}} area corresponds to smaller attention weight. \textbf{Left:} The attention weights of continuous tokens to the 1024 scene tokens in the vanilla continuous-action-based policy. 
    \textbf{Middle:} The attention weights of point flow tokens to the scene tokens in \modelname. \textbf{Right:} The attention weights of keypose tokens in \modelname.}
    \label{fig:general-visual}
\end{figure}

As shown in Figure~\ref{fig:general-visual}, we visualize attention weights for our policy and the vanilla continuous-based policy in the "Lift Tray" task on RLBench2. 
The left image reveals that in the continuous-based policy, action tokens fail to consistently focus on the tray—the key task-relevant object and disperse attention across both the tray and the robotic arm. 
This suggests why continuous-based policies often struggle with precise localization, as observed in previous experiments.

In contrast, as shown in the middle and right images, \modelname's pointflow and keypose tokens strongly attend to the tray, with pointflow tokens specifically concentrating on its edges—where the gripper is poised to grasp. 
Since the initial query points are sampled from the tray rather than the cube atop it, the model learns to deprioritize the cube. 
This explains why in the "Carry the Tray" task, replacing in-distribution cups with out-of-distribution Rubik’s Cubes and colorful cubes does not impair \modelname’s localization performance. 
By strategically selecting initial query points during training, the model learns to track the overall object's motion, improving generalization under disturbances.

Further, we examine how these interfaces guide continuous action prediction. 
As shown in Figure~\ref{fig:general-heatmap}, continuous action tokens exhibit significantly higher attention weights toward keypose and pointflow tokens than scene tokens, underscoring the critical role of interfaces in action prediction. 
By incorporating target gripper poses and object pointflow as interfaces, \modelname not only maintains focus on task-relevant regions despite scene variations but also alleviates the learning burden on action tokens by distilling key information, rather than directly querying from the entire scene.

\section{Conclusion and Limitation} 
\label{sec:conclusion}
We introduce \modelname, an end-to-end interface-based manipulation policy that leverages target gripper poses and object pointflow. 
\modelname achieves state-of-the-art performance on the RLBench2 simulation benchmark and demonstrates strong effectiveness and robustness in real-world experiments.

\textbf{Limitation.} There remain two main limitations.
First, the computational cost of visual foundation models and the diffusion process constrains efficiency. 
Future work will focus on accelerating diffusion sampling and adopting lightweight vision models.
Second, cross-embodiment evaluation on different robotic platforms is essential to assess \modelname's generalization across hardware.



\bibliographystyle{plainnat}
\bibliography{short,references}

\begin{thebibliography}{49}
\providecommand{\natexlab}[1]{#1}
\providecommand{\url}[1]{\texttt{#1}}
\expandafter\ifx\csname urlstyle\endcsname\relax
  \providecommand{\doi}[1]{doi: #1}\else
  \providecommand{\doi}{doi: \begingroup \urlstyle{rm}\Url}\fi

\bibitem[Bharadhwaj et~al.(2024)Bharadhwaj, Mottaghi, Gupta, and Tulsiani]{track2act}
Homanga Bharadhwaj, Roozbeh Mottaghi, Abhinav Gupta, and Shubham Tulsiani.
\newblock Track2act: Predicting point tracks from internet videos enables generalizable robot manipulation.
\newblock In \emph{Arxiv}, 2024.

\bibitem[Bowen et~al.(2024)Bowen, Wei, Jan, and Stan]{foundationpose}
Wen Bowen, Yang Wei, Kautz Jan, and Birchfield Stan.
\newblock {FoundationPose}: Unified 6d pose estimation and tracking of novel objects.
\newblock In \emph{CVPR}, 2024.

\bibitem[Brohan et~al.(2022)Brohan, Brown, Carbajal, Chebotar, Dabis, Finn, Gopalakrishnan, Hausman, Herzog, Hsu, et~al.]{rt1}
Anthony Brohan, Noah Brown, Justice Carbajal, Yevgen Chebotar, Joseph Dabis, Chelsea Finn, Keerthana Gopalakrishnan, Karol Hausman, Alex Herzog, Jasmine Hsu, et~al.
\newblock Rt-1: Robotics transformer for real-world control at scale.
\newblock In \emph{Arxiv}, 2022.

\bibitem[Chi et~al.(2023)Chi, Xu, Feng, Cousineau, Du, Burchfiel, Tedrake, and Song]{dp}
Cheng Chi, Zhenjia Xu, Siyuan Feng, Eric Cousineau, Yilun Du, Benjamin Burchfiel, Russ Tedrake, and Shuran Song.
\newblock Diffusion policy: Visuomotor policy learning via action diffusion.
\newblock In \emph{RSS}, 2023.

\bibitem[Devgon et~al.(2020)Devgon, Ichnowski, Balakrishna, Zhang, and Goldberg]{orientpose}
Shivin Devgon, Jeffrey Ichnowski, Ashwin Balakrishna, Harry Zhang, and Ken Goldberg.
\newblock Orienting novel 3d objects using self-supervised learning of rotation transforms.
\newblock In \emph{CASE}, 2020.

\bibitem[Driess et~al.(2023)Driess, Xia, Sajjadi, Lynch, Chowdhery, Ichter, Wahid, Tompson, Vuong, Yu, et~al.]{palm-e}
Danny Driess, Fei Xia, Mehdi~SM Sajjadi, Corey Lynch, Aakanksha Chowdhery, Brian Ichter, Ayzaan Wahid, Jonathan Tompson, Quan Vuong, Tianhe Yu, et~al.
\newblock Palm-e: An embodied multimodal language model.
\newblock In \emph{Arxiv}, 2023.

\bibitem[Eisner et~al.(2022)Eisner, Zhang, and Held]{flowbot3d}
Ben Eisner, Harry Zhang, and David Held.
\newblock Flowbot3d: Learning 3d articulation flow to manipulate articulated objects.
\newblock In \emph{RSS}, 2022.

\bibitem[Fang et~al.(2023)Fang, Wang, Fang, Gou, Liu, Yan, Liu, Xie, and Lu]{anygrasp}
Hao-Shu Fang, Chenxi Wang, Hongjie Fang, Minghao Gou, Jirong Liu, Hengxu Yan, Wenhai Liu, Yichen Xie, and Cewu Lu.
\newblock Anygrasp: Robust and efficient grasp perception in spatial and temporal domains.
\newblock In \emph{TRO}, 2023.

\bibitem[Grotz et~al.(2024)Grotz, Shridhar, Asfour, and Fox]{peract2}
Markus Grotz, Mohit Shridhar, Tamim Asfour, and Dieter Fox.
\newblock Peract2: A perceiver actor framework for bimanual manipulation tasks.
\newblock In \emph{Arxiv}, 2024.

\bibitem[Ho et~al.(2020)Ho, Jain, and Abbeel]{ddpm}
Jonathan Ho, Ajay Jain, and Pieter Abbeel.
\newblock Denoising diffusion probabilistic models.
\newblock In \emph{NeurIPS}, 2020.

\bibitem[James and Davison(2022)]{qattention}
Stephen James and Andrew~J. Davison.
\newblock Q-attention: Enabling efficient learning for vision-based robotic manipulation.
\newblock In \emph{IEEE Robotics and Automation Letters}, 2022.

\bibitem[Karaev et~al.(2024)Karaev, Rocco, Graham, Neverova, Vedaldi, and Rupprecht]{cotracker}
Nikita Karaev, Ignacio Rocco, Benjamin Graham, Natalia Neverova, Andrea Vedaldi, and Christian Rupprecht.
\newblock Cotracker: It is better to track together.
\newblock In \emph{ECCV}, 2024.

\bibitem[Ke et~al.(2024)Ke, Gkanatsios, and Fragkiadaki]{3d-diffuser-actor}
Tsung-Wei Ke, Nikolaos Gkanatsios, and Katerina Fragkiadaki.
\newblock 3d diffuser actor: Policy diffusion with 3d scene representations.
\newblock In \emph{CoRL}, 2024.

\bibitem[Kim et~al.(2024)Kim, Pertsch, Karamcheti, Xiao, Balakrishna, Nair, Rafailov, Foster, Lam, Sanketi, et~al.]{openvla}
Moo~Jin Kim, Karl Pertsch, Siddharth Karamcheti, Ted Xiao, Ashwin Balakrishna, Suraj Nair, Rafael Rafailov, Ethan Foster, Grace Lam, Pannag Sanketi, et~al.
\newblock Openvla: An open-source vision-language-action model.
\newblock In \emph{CoRL}, 2024.

\bibitem[Kirillov et~al.(2023)Kirillov, Mintun, Ravi, Mao, Rolland, Gustafson, Xiao, Whitehead, Berg, Lo, Doll{\'a}r, and Girshick]{sam}
Alexander Kirillov, Eric Mintun, Nikhila Ravi, Hanzi Mao, Chloe Rolland, Laura Gustafson, Tete Xiao, Spencer Whitehead, Alexander~C. Berg, Wan-Yen Lo, Piotr Doll{\'a}r, and Ross Girshick.
\newblock Segment anything.
\newblock In \emph{Arxiv}, 2023.

\bibitem[Liu et~al.(2024{\natexlab{a}})Liu, Arthur, He, Seita, and Sukhatme]{voxact-b}
I~Liu, Chun Arthur, Sicheng He, Daniel Seita, and Gaurav Sukhatme.
\newblock Voxact-b: Voxel-based acting and stabilizing policy for bimanual manipulation.
\newblock In \emph{CoRL}, 2024{\natexlab{a}}.

\bibitem[Liu et~al.(2023)Liu, Zeng, Ren, Li, Zhang, Yang, Li, Yang, Su, Zhu, et~al.]{groundingdino}
Shilong Liu, Zhaoyang Zeng, Tianhe Ren, Feng Li, Hao Zhang, Jie Yang, Chunyuan Li, Jianwei Yang, Hang Su, Jun Zhu, et~al.
\newblock Grounding dino: Marrying dino with grounded pre-training for open-set object detection.
\newblock In \emph{Arxiv}, 2023.

\bibitem[Liu et~al.(2024{\natexlab{b}})Liu, Wu, Li, Tan, Chen, Wang, Xu, Su, and Zhu]{rdt}
Songming Liu, Lingxuan Wu, Bangguo Li, Hengkai Tan, Huayu Chen, Zhengyi Wang, Ke~Xu, Hang Su, and Jun Zhu.
\newblock Rdt-1b: a diffusion foundation model for bimanual manipulation.
\newblock In \emph{Arxiv}, 2024{\natexlab{b}}.

\bibitem[Oquab et~al.(2023)Oquab, Darcet, Moutakanni, Vo, Szafraniec, Khalidov, Fernandez, Haziza, Massa, El-Nouby, Howes, Huang, Xu, Sharma, Li, Galuba, Rabbat, Assran, Ballas, Synnaeve, Misra, Jegou, Mairal, Labatut, Joulin, and Bojanowski]{dinov2}
Maxime Oquab, Timothée Darcet, Theo Moutakanni, Huy~V. Vo, Marc Szafraniec, Vasil Khalidov, Pierre Fernandez, Daniel Haziza, Francisco Massa, Alaaeldin El-Nouby, Russell Howes, Po-Yao Huang, Hu~Xu, Vasu Sharma, Shang-Wen Li, Wojciech Galuba, Mike Rabbat, Mido Assran, Nicolas Ballas, Gabriel Synnaeve, Ishan Misra, Herve Jegou, Julien Mairal, Patrick Labatut, Armand Joulin, and Piotr Bojanowski.
\newblock Dinov2: Learning robust visual features without supervision.
\newblock In \emph{Arxiv}, 2023.

\bibitem[Pan et~al.(2023)Pan, Okorn, Zhang, Eisner, and Held]{taxpose}
Chuer Pan, Brian Okorn, Harry Zhang, Ben Eisner, and David Held.
\newblock Tax-pose: Task-specific cross-pose estimation for robot manipulation.
\newblock In \emph{CoRL}, 2023.

\bibitem[Perez et~al.(2017)Perez, Strub, de~Vries, Dumoulin, and Courville]{film}
Ethan Perez, Florian Strub, Harm de~Vries, Vincent Dumoulin, and Aaron Courville.
\newblock Film: Visual reasoning with a general conditioning layer.
\newblock In \emph{Proceedings of the AAAI Conference on Artificial Intelligence}, 2017.

\bibitem[Qi et~al.(2024)Qi, Wu, Yu, Liu, Jiang, Lin, and Held]{tooluseflow}
Carl Qi, Yilin Wu, Lifan Yu, Haoyue Liu, Bowen Jiang, Xingyu Lin, and David Held.
\newblock Learning generalizable tool-use skills through trajectory generation.
\newblock In \emph{IROS}, 2024.

\bibitem[Qi et~al.(2017)Qi, Yi, Su, and Guibas]{pointnet}
Charles~R. Qi, Li~Yi, Hao Su, and Leonidas~J. Guibas.
\newblock Pointnet++: Deep hierarchical feature learning on point sets in a metric space.
\newblock In \emph{Arxiv}, 2017.

\bibitem[Radford et~al.(2021)Radford, Kim, Hallacy, Ramesh, Goh, Agarwal, Sastry, Askell, Mishkin, Clark, Krueger, and Sutskever]{clip}
Alec Radford, Jong~Wook Kim, Chris Hallacy, Aditya Ramesh, Gabriel Goh, Sandhini Agarwal, Girish Sastry, Amanda Askell, Pamela Mishkin, Jack Clark, Gretchen Krueger, and Ilya Sutskever.
\newblock Learning transferable visual models from natural language supervision.
\newblock In \emph{Arxiv}, 2021.

\bibitem[Ravi et~al.(2024)Ravi, Gabeur, Hu, Hu, Ryali, Ma, Khedr, R{\"a}dle, Rolland, Gustafson, et~al.]{sam2}
Nikhila Ravi, Valentin Gabeur, Yuan-Ting Hu, Ronghang Hu, Chaitanya Ryali, Tengyu Ma, Haitham Khedr, Roman R{\"a}dle, Chloe Rolland, Laura Gustafson, et~al.
\newblock Sam 2: Segment anything in images and videos.
\newblock In \emph{Arxiv}, 2024.

\bibitem[Seita et~al.(2023)Seita, Wang, Shetty, Li, Erickson, and Held]{toolflownet}
Daniel Seita, Yufei Wang, Sarthak~J Shetty, Edward~Yao Li, Zackory Erickson, and David Held.
\newblock Toolflownet: Robotic manipulation with tools via predicting tool flow from point clouds.
\newblock In \emph{CoRL}, 2023.

\bibitem[Shridhar et~al.(2022)Shridhar, Manuelli, and Fox]{perceiver}
Mohit Shridhar, Lucas Manuelli, and Dieter Fox.
\newblock Perceiver-actor: A multi-task transformer for robotic manipulation.
\newblock In \emph{CoRL}, 2022.

\bibitem[Song et~al.(2020)Song, Meng, and Ermon]{ddim}
Jiaming Song, Chenlin Meng, and Stefano Ermon.
\newblock Denoising diffusion implicit models.
\newblock In \emph{Arxiv}, 2020.

\bibitem[Su et~al.(2023)Su, Lu, Pan, Murtadha, Wen, and Liu]{rotary}
Jianlin Su, Yu~Lu, Shengfeng Pan, Ahmed Murtadha, Bo~Wen, and Yunfeng Liu.
\newblock Roformer: Enhanced transformer with rotary position embedding.
\newblock In \emph{Arxiv}, 2023.

\bibitem[Sucan et~al.(2012)Sucan, Moll, and Kavraki]{mplib}
Ioan~A Sucan, Mark Moll, and Lydia~E Kavraki.
\newblock The open motion planning library.
\newblock In \emph{IEEE Robotics \& Automation Magazine}, 2012.

\bibitem[Sundaralingam et~al.(2023)Sundaralingam, Hari, Fishman, Garrett, Van~Wyk, Blukis, Millane, Oleynikova, Handa, Ramos, et~al.]{curobo}
Balakumar Sundaralingam, Siva Kumar~Sastry Hari, Adam Fishman, Caelan Garrett, Karl Van~Wyk, Valts Blukis, Alexander Millane, Helen Oleynikova, Ankur Handa, Fabio Ramos, et~al.
\newblock Curobo: Parallelized collision-free robot motion generation.
\newblock In \emph{ICRA}, 2023.

\bibitem[Tian et~al.(2023)Tian, Zhang, Yin, and Dong]{sgtapose}
Yang Tian, Jiyao Zhang, Zekai Yin, and Hao Dong.
\newblock Robot structure prior guided temporal attention for camera-to-robot pose estimation from image sequence.
\newblock In \emph{CVPR}, 2023.

\bibitem[Tian et~al.(2024)Tian, Zhang, Huang, Wang, Wang, Pang, and Dong]{robokeygen}
Yang Tian, Jiyao Zhang, Guowei Huang, Bin Wang, Ping Wang, Jiangmiao Pang, and Hao Dong.
\newblock Robokeygen: Robot pose and joint angles estimation via diffusion-based 3d keypoint generation.
\newblock In \emph{ICRA}, 2024.

\bibitem[Tian et~al.(2025)Tian, Yang, Zeng, Wang, Lin, Dong, and Pang]{seer}
Yang Tian, Sizhe Yang, Jia Zeng, Ping Wang, Dahua Lin, Hao Dong, and Jiangmiao Pang.
\newblock Predictive inverse dynamics models are scalable learners for robotic manipulation.
\newblock In \emph{ICLR}, 2025.

\bibitem[Wang et~al.(2024{\natexlab{a}})Wang, Fang, Fang, and Lu]{rise}
Chenxi Wang, Hongjie Fang, Hao-Shu Fang, and Cewu Lu.
\newblock Rise: 3d perception makes real-world robot imitation simple and effective.
\newblock In \emph{IROS}, 2024{\natexlab{a}}.

\bibitem[Wang et~al.(2024{\natexlab{b}})Wang, Chen, Huang, Ben, Wang, Mi, Huang, Zhao, Chen, Yang, et~al.]{grutopia}
Hanqing Wang, Jiahe Chen, Wensi Huang, Qingwei Ben, Tai Wang, Boyu Mi, Tao Huang, Siheng Zhao, Yilun Chen, Sizhe Yang, et~al.
\newblock Grutopia: Dream general robots in a city at scale.
\newblock In \emph{Arxiv}, 2024{\natexlab{b}}.

\bibitem[Wang et~al.(2024{\natexlab{c}})Wang, Yin, Huang, Kelestemur, Wang, and Li]{gendp}
Yixuan Wang, Guang Yin, Binghao Huang, Tarik Kelestemur, Jiuguang Wang, and Yunzhu Li.
\newblock Gendp: 3d semantic fields for category-level generalizable diffusion policy.
\newblock In \emph{CoRL}, 2024{\natexlab{c}}.

\bibitem[Wang et~al.(2024{\natexlab{d}})Wang, Zhang, Li, Kelestemur, Driggs-Campbell, Wu, Fei-Fei, and Li]{d3fields}
Yixuan Wang, Mingtong Zhang, Zhuoran Li, Tarik Kelestemur, Katherine Driggs-Campbell, Jiajun Wu, Li~Fei-Fei, and Yunzhu Li.
\newblock D$^3$fields: Dynamic 3d descriptor fields for zero-shot generalizable rearrangement.
\newblock In \emph{CoRL}, 2024{\natexlab{d}}.

\bibitem[Wen et~al.(2023)Wen, Tremblay, Blukis, Tyree, M{\"u}ller, Evans, Fox, Kautz, and Birchfield]{bundlesdf}
Bowen Wen, Jonathan Tremblay, Valts Blukis, Stephen Tyree, Thomas M{\"u}ller, Alex Evans, Dieter Fox, Jan Kautz, and Stan Birchfield.
\newblock Bundlesdf: Neural 6-dof tracking and 3d reconstruction of unknown objects.
\newblock In \emph{Proceedings of the IEEE/CVF Conference on Computer Vision and Pattern Recognition}, pages 606--617, 2023.

\bibitem[Wen et~al.(2024)Wen, Lin, So, Chen, Dou, Gao, and Abbeel]{atm}
Chuan Wen, Xingyu Lin, John So, Kai Chen, Qi~Dou, Yang Gao, and Pieter Abbeel.
\newblock Any-point trajectory modeling for policy learning.
\newblock In \emph{RSS}, 2024.

\bibitem[Weng et~al.(2022)Weng, Bajracharya, Wang, Agrawal, and Held]{fabricflownet}
Thomas Weng, Sujay~Man Bajracharya, Yufei Wang, Khush Agrawal, and David Held.
\newblock Fabricflownet: Bimanual cloth manipulation with a flow-based policy.
\newblock In \emph{CoRL}, 2022.

\bibitem[Wu et~al.(2024)Wu, Shentu, Yi, Lin, and Abbeel]{wu2024gello}
Philipp Wu, Yide Shentu, Zhongke Yi, Xingyu Lin, and Pieter Abbeel.
\newblock Gello: A general, low-cost, and intuitive teleoperation framework for robot manipulators.
\newblock In \emph{IROS}, 2024.

\bibitem[Xu et~al.(2024)Xu, Xu, Xu, Chi, Wetzstein, Veloso, and Song]{im2flow2act}
Mengda Xu, Zhenjia Xu, Yinghao Xu, Cheng Chi, Gordon Wetzstein, Manuela Veloso, and Shuran Song.
\newblock Flow as the cross-domain manipulation interface.
\newblock In \emph{CoRL}, 2024.

\bibitem[Yu et~al.(2024)Yu, Xu, Chen, Ren, and Pan]{bikc}
Dongjie Yu, Hang Xu, Yizhou Chen, Yi~Ren, and Jia Pan.
\newblock Bikc: Keypose-conditioned consistency policy for bimanual robotic manipulation.
\newblock In \emph{Arxiv}, 2024.

\bibitem[Yuan et~al.(2024)Yuan, Wen, Zhang, and Gao]{generalflow}
Chengbo Yuan, Chuan Wen, Tong Zhang, and Yang Gao.
\newblock General flow as foundation affordance for scalable robot learning.
\newblock In \emph{CoRL}, 2024.

\bibitem[Ze et~al.(2024)Ze, Zhang, Zhang, Hu, Wang, and Xu]{dp3}
Yanjie Ze, Gu~Zhang, Kangning Zhang, Chenyuan Hu, Muhan Wang, and Huazhe Xu.
\newblock 3d diffusion policy.
\newblock In \emph{RSS}, 2024.

\bibitem[Zhang et~al.(2023)Zhang, Eisner, and Held]{flowbot++}
Harry Zhang, Ben Eisner, and David Held.
\newblock Flowbot++: Learning generalized articulated objects manipulation via articulation projection.
\newblock In \emph{CoRL}, 2023.

\bibitem[Zhao et~al.(2023{\natexlab{a}})Zhao, Kumar, Levine, and Finn]{act}
Tony~Z Zhao, Vikash Kumar, Sergey Levine, and Chelsea Finn.
\newblock Learning fine-grained bimanual manipulation with low-cost hardware.
\newblock In \emph{RSS}, 2023{\natexlab{a}}.

\bibitem[Zhao et~al.(2023{\natexlab{b}})Zhao, Wu, Chen, Zhang, Fan, Mo, and Dong]{dualafford}
Yan Zhao, Ruihai Wu, Zhehuan Chen, Yourong Zhang, Qingnan Fan, Kaichun Mo, and Hao Dong.
\newblock Dualafford: Learning collaborative visual affordance for dual-gripper manipulation.
\newblock In \emph{ICLR}, 2023{\natexlab{b}}.

\end{thebibliography}

\clearpage

\appendix
\subsection{Foundation Models}
\modelname leverages foundation models in various aspects, either to generate ground-truth labels or serve as encoders. In this section, we will specifically describe how we utilize these foundation models, along with the exact versions and settings of the models we use.

\textbf{3D Semantic Fields.} To obtain semantic features from 2D images, we leverage the DINOv2 ViT-S/14 model, with a feature dimension of 384. In practice, we crop and downsample the scene point clouds to 6144 in simulations and 3072 in real-world experiments, resulting in 3D semantic fields with dimensions of 6144×387 and 3072×387, respectively, which will be further downsampled by PointNet++ in the subsequent processing. Here, the dimension of 387 includes both the semantic features and the xyz coordinates of the points.

\textbf{Initial Point Sampling.} In both simulation and real-world experiments, we sample 200 points on the object, which has been shown to effectively capture the object's motion. For simulation tasks, we typically select a single camera from which the target object is visible at the first timestep in each episode. We then feed a language prompt to the SwinB CogCoor version of the Grounding DINO model to obtain the bounding box for the target object from that camera. This bounding box is subsequently passed to the ViT-B SAM model, which generates a mask for the pixels corresponding to the target object.

\textbf{Language.} Text instructions are encoded using the CLIP ResNet50 text encoder and projected through a linear layer to generate the language token.

\textbf{Object Pose Estimation.}
In simulation, we generate a dataset that includes the ground truth of 6D pose of the objects.
During the real-world data collection process, we utilize BundleSDF\cite{bundlesdf} to obtain the meshes of unknown objects with a coarse 6D pose.
With the meshes of objects, we employ the model-based method of FoundationPose\cite{foundationpose} to enhance the precision of the objects' 6D pose.
More specifically, we perform 6D pose estimation on the first frame of each demonstration, and for each subsequent frame, we use 6D pose tracking.

\subsection{Architecture Details}

\textbf{Relative Attention.} Our 3D denoising transformer takes four types of 3D tokens as input: point flow tokens \( F \), keyframe action tokens \( a_t^{k,i} \), continuous action tokens \( a_t^{c,i} \), and scene point cloud tokens \( S_t \). Each 3D token is represented by a latent embedding and a 3D position. Following the 3D Diffuser Actor \cite{3d-diffuser-actor}, our model applies relative self-attention among all 3D tokens. We use a rotary positional embedding to encode relative positional information in the attention layers, with the attention weight between query \( q \) and key \( k \) given by: 
\[
e_{q,k} \propto x_q^T M(c_q - c_k) x_k
\]
where \( x_q \) and \( x_k \) are the features of the query and key, and \( c_q \) and \( c_k \) are the corresponding 3D coordinates of the query and key tokens. \( M \) is a matrix function that depends only on the relative positions of the 3D coordinates of the query and key tokens. Relative attention allows us to effectively leverage the relative 3D spatial information between tokens within the transformer.

\textbf{Robot State.} For each gripper, the arm state consists of the end-effector position, quaternion, and gripper state, forming an 8-dimensional vector. The gripper state is either 0 (closed) or 1 (open). In our 3d denoising transformer, our method uses separate tokens for the left and right hand actions, helping the model focus on different regions. For predicting 50 continuous actions, we use 100 tokens, and for predicting 4 keyframe actions, we use 8 tokens. The robot state is tokenized with a three-layer MLP, similar to how we tokenize language and initial query points.

\begin{figure}[htbp]
    \centering
    \includegraphics[width=1\linewidth]{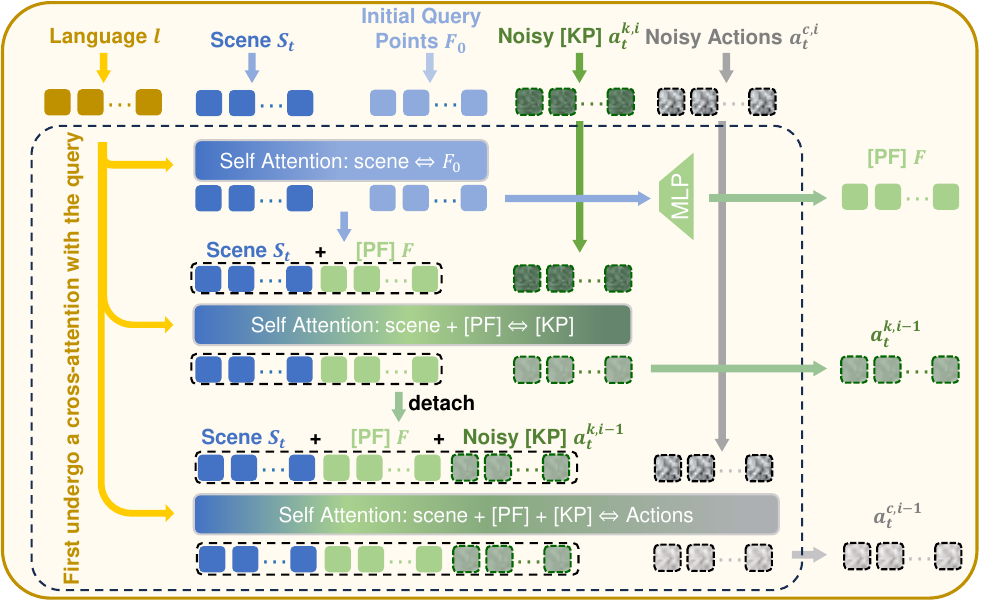}
    \caption{\textbf{Detailed architecture of 3D denoising model.}}
    \label{fig:model-detail}
\end{figure}

\textbf{Network Architecture.} Next, we detail the structure of our 3D denoising model as shown in Figure~\ref{fig:model-detail}. At timestep \( t \) and denoising step \( i \), let \( a_t^{k,i} \) and \( a_t^{c,i} \) be the noised keyframe action \( a_t^{k} \) and continuous action \( a_t^{c} \), respectively. The transformer incorporates the scene tokens \( S_t \), language tokens \( l \), query points tokens \( F_0 \), and the noised action tokens \( a_t^{k,i} \) and \( a_t^{c,i} \).

All tokens in each self-attention block will first perform parallel attention with the language tokens \( l \) to capture the relevant language instructions.
Next, we concatenate the query points tokens \( F_0 \) and scene tokens \( S_t \), applying relative self-attention. The resulting tokens for \( F_0 \) pass through three MLP layers, where each token outputs a 12-dimensional vector representing the xyz coordinates of that point for the next 4 keyframes. Then, we perform relative self-attention on the noised keyframe action tokens \( a_t^{k,i} \) and the pointflow tokens and the scene tokens from the previous layer. The output corresponding to the keypose token is passed through three MLP layers to map it to the action dimension, with supervision provided by the keypose noise \( \epsilon_t^{k,i} \). Finally, we perform relative self-attention between the noised continuous action tokens \( a_t^{c,i} \) and the detached vector from the previous layer. The output corresponding to the continuous token is also passed through three MLP layers to map it to the action dimension, with supervision provided by the continuous noise \( \epsilon_t^{c,i} \).

\subsection{Simple PPI}
In the real-world, the original PPI has an average inference speed of 2.63 Hz.
Since the predicted actions are dense and continuous, we excute the first few steps (e.g., 3 steps) of the 50 predicted actions during inference, which maintains performace while ensuring smooth motion.
Additionally, we implemented Simple PPI to further reduce computation costs.
Simple PPI achieves 10.52 Hz by reducing whole scene tokens (3072 → 512), downsampled scene tokens (512 → 64), pointflow tokens (200 → 50), DDIM denoising steps (20 → 5), and halving the number of transformer layers in real-world settings. 
As shown in Table~\ref{Performance of Simple PPI}, this results in nearly 4× faster inference without much accuracy loss.

\begin{table}[htbp]
\centering
\caption{\textbf{Performance of Simple PPI.}}
\label{Performance of Simple PPI}
\resizebox{\linewidth}{!}{\begin{tabular}{cccc}
\toprule
\multirow{2}{*}{Method} & 
\multicolumn{3}{c}{SR (\%) $\uparrow$ /Loc-SR (\%) $\uparrow$ / Normalized Score $\uparrow$} \\
\cline{2-4}
& Carry the Tray
& Handover \& Insert the Plate
& Wipe the Plate
\\
\midrule \midrule 
Simple PPI (10.52 Hz)
& 50.0 / 70.0 / 6.0
& 30.0 / 100 / 5.7
& 50.0 / 70.0 / 6.8
\\
\baseline{PPI (2.63 Hz)} 
& \baseline{\textbf{50.0} / \textbf{100} / \textbf{7.8}}
& \baseline{\textbf{40.0} / \textbf{100} / \textbf{7.7}}
& \baseline{\textbf{70.0} / \textbf{70.0} / \textbf{8.3}}
\\
\bottomrule
\end{tabular}}
\end{table}

\subsection{More Tasks in Real-World}
To demonstrate PPI’s ability to handle high-precision tasks and manipulate deformable objects, we design two new real-world tasks.  
For high precision task \textbf{Press the Cap of Bottle}, one arm picks up a bottle and places it on a 3cm × 3cm cube, while the other precisely presses its cap (0.75cm radius). 
For deformable object task \textbf{Wear the Scarf}, arms are required to lift a scarf simultaneously and wrap it around a mannequin, where CoTracker is used to obtain pointflow ground truth. 
To showcase PPI's capability to align the scanner and bottle from different initial states and complete single-arm tasks, we design a single-arm experiment \textbf{Scan the Bottle with Single Arm} where the bottle starts in two directions: 1) facing the scanner and 2) at a 45-degree angle, requiring the scanner to rotate for a successful scan.  
The procedures for these three tasks are available on \href{https://yuyinyang3y.github.io/PPI/}{https://yuyinyang3y.github.io/PPI/}.
As shown in Table~\ref{new-tasks}, our method outperforms the baseline (ACT) across all tasks. 
\begin{table}[htbp]
\centering
\caption{\textbf{Real-world results of three extra tasks.}}
\label{new-tasks}
\resizebox{\linewidth}{!}{\begin{tabular}{cccc}
\toprule
\multirow{2}{*}{Method} & 
\multicolumn{3}{c}{SR (\%) $\uparrow$ /Loc-SR (\%) $\uparrow$ / Normalized Score $\uparrow$} \\
\cline{2-4}
& Wear the Scarf
& Press the Cap of Bottle
& Scan the Bottle (Unimanual)
\\
\midrule \midrule 
ACT 
& 40.0 / 70.0 / 5.7
& 10.0 / 20.0 / 3.0
& 20.0 / 70.0 / 4.7
\\
\baseline{Ours} 
& \baseline{\textbf{70.0} / \textbf{80.0} / \textbf{7.7}}
& \baseline{\textbf{40.0} / \textbf{80.0} / \textbf{6.7}}
& \baseline{\textbf{80.0} / \textbf{100} / \textbf{9.3}}
\\
\bottomrule
\end{tabular}}
\end{table}

\subsection{Baseline Implementation}
In the simulation benchmark, we report the scores for DP3, ACT, PerAct$^2$, and 3D Diffuser Actor from their respective papers. For ACT and PerAct$^2$, we reproduce the results using the official code in RLBench2 and train each method for 100k steps, as suggusted in the original paper. The final scores are obtained by averaging the scores from epochs 99900, 99800, and 99700. For DP3, we directly double its action dimension and adapt it to the bimanual task setting. We use the default hyperparameter settings, prediction horizon, observation step, and other configurations from the paper, and train the model for 3000 epochs as suggested. We average the success rate of epoch 3000, 2900 and 2800 for each task. Similar to our own method, we double the number of action tokens for 3D Diffuser Actor, using separate actions to predict the left and right hand movements instead of directly doubling the action dimension. We select the checkpoint corresponding to the epoch when the eval location and rotation accuracy converged during the model training process. For simple tasks like ball lifting, we chose steps 10000, 11000, and 12000, while for more complex tasks such as handover easy, we select steps 16000, 18000, and 20000.

For real-world experiments, we implement the baselines in the same way as in simulations, but trained them with different durations due to the increased complexity of real-world tasks: ACT for 10000 epochs, DP3 for 8000 epochs (with its prediction horizon $N_{act}$ adjusted to 10 for better performance), and 3D Diffuser Actor for 100000 steps.

\subsection{Detailed Real-world Experiment Settings}
\begin{figure}[htbp]
    \centering
    \includegraphics[width=0.8\linewidth]{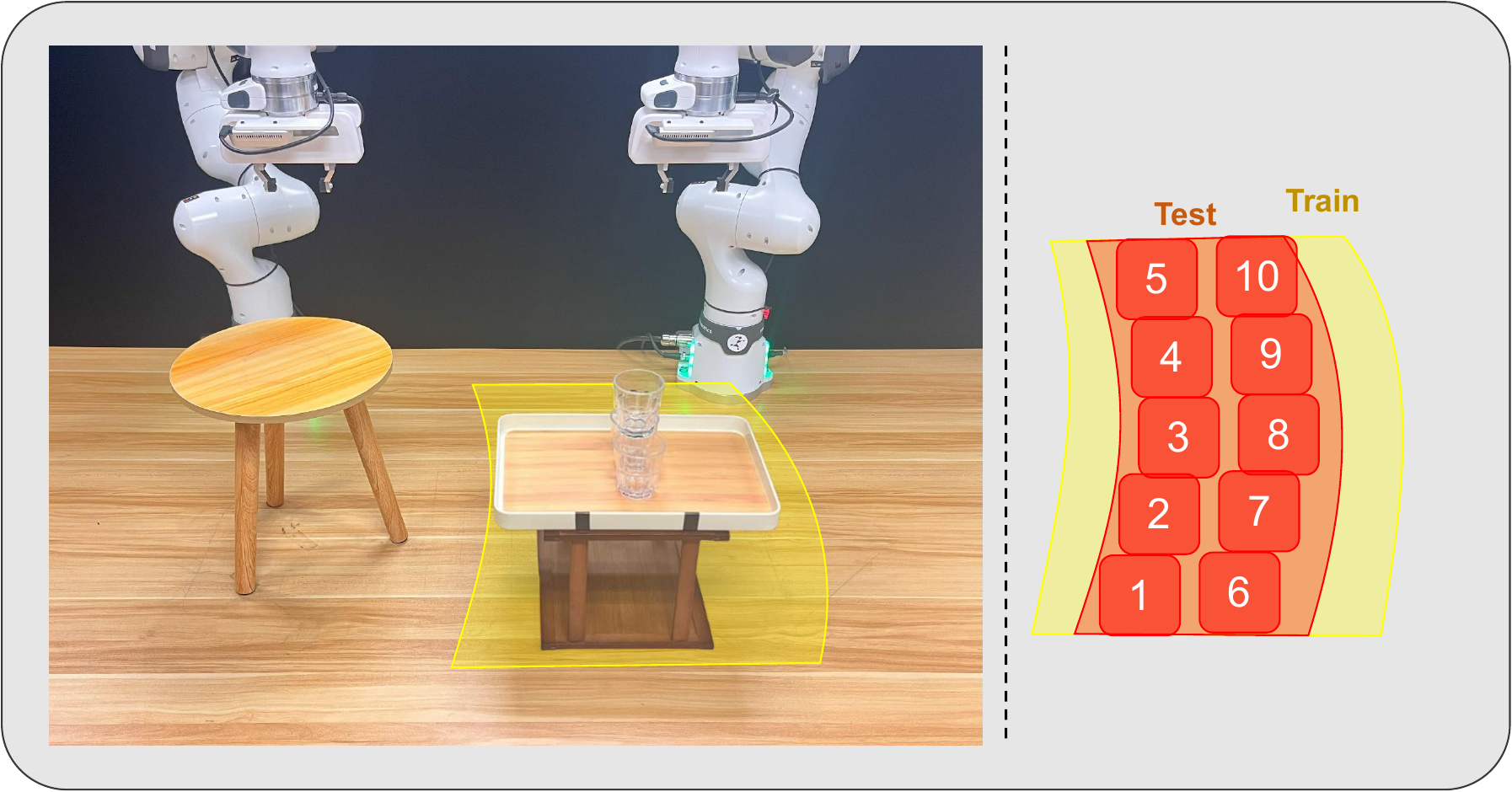}
    \caption{
    \textbf{Detailed 10 settings of Carry the Tray.}
    The tray remains relatively stationary while the platform is moved to 10 different locations, serving as 10 distinct settings.
    }
    \label{fig:exp-carry}
\end{figure}

\begin{figure}[htbp]
    \centering
    \includegraphics[width=0.8\linewidth]{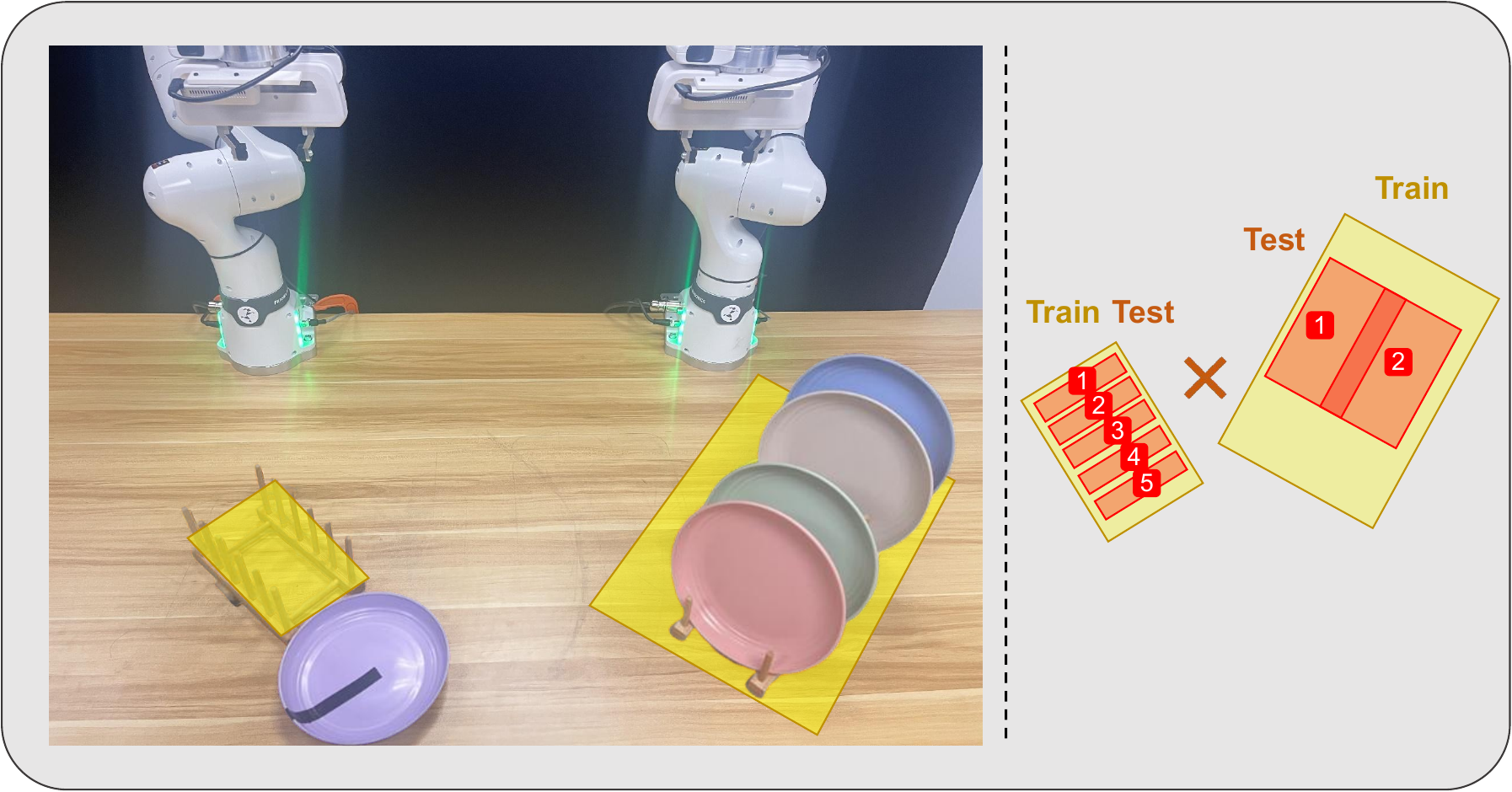}
    \caption{
    \textbf{Detailed 10 settings of Handover and Insert the Plate.}
    On the left, the plate varies across 5 slots on the rack, while on the right, the overall setup shifts between 2 designated positions, combining to form 10 different settings.
    }
    \label{fig:exp-hand}
\end{figure}

\begin{figure}[htbp]
    \centering
    \includegraphics[width=0.8\linewidth]{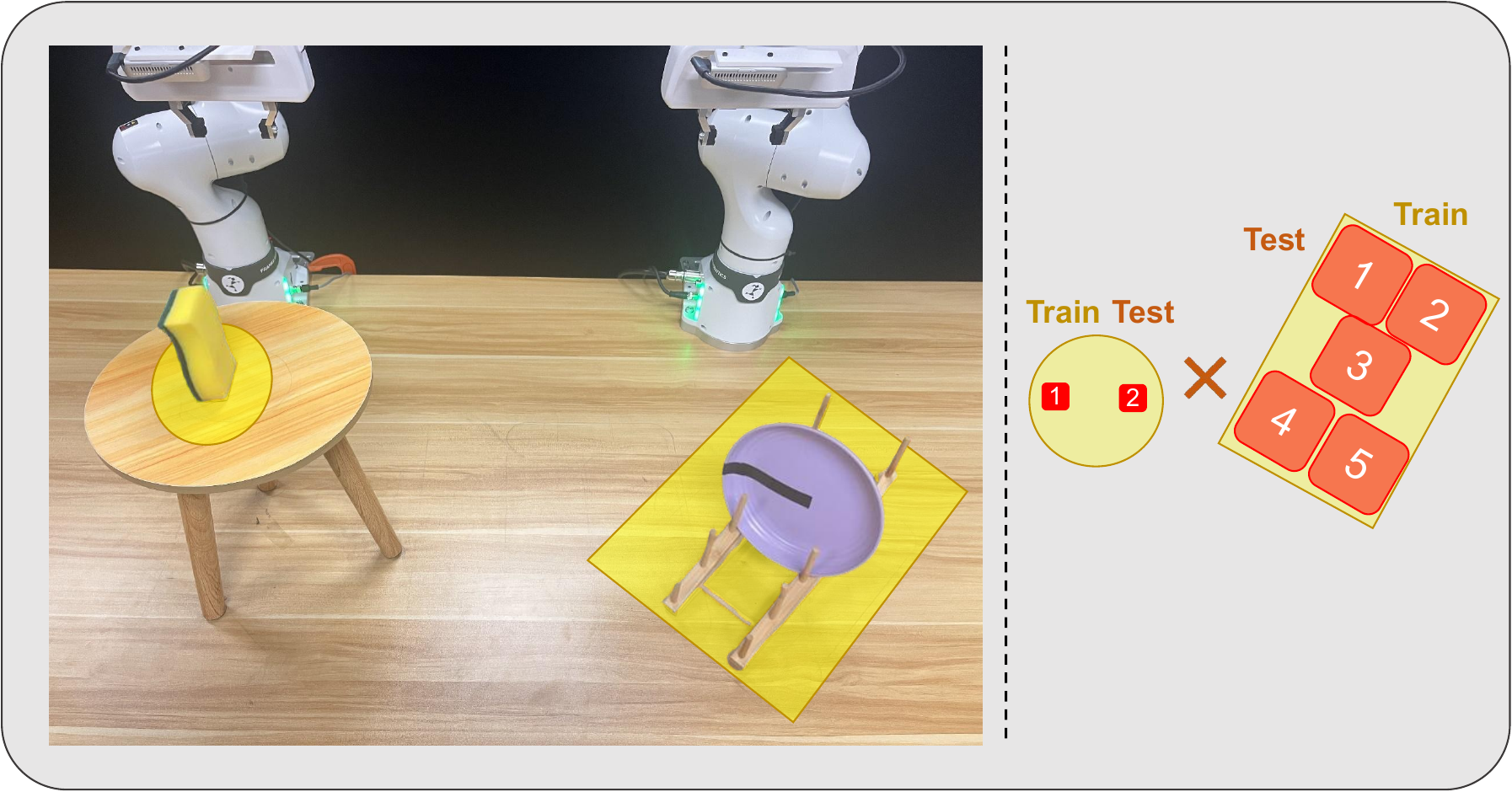}
    \caption{
    \textbf{Detailed 10 settings of Wipe the Plate.}
    On the left, the sponge changes positions between the 2 sides of the yellow training area, and on the right, the plate moves across the 5 red areas shown, creating a total of 10 settings.
    }
    \label{fig:exp-wipe}
\end{figure}

\begin{figure}[htbp]
    \centering
    \includegraphics[width=0.8\linewidth]{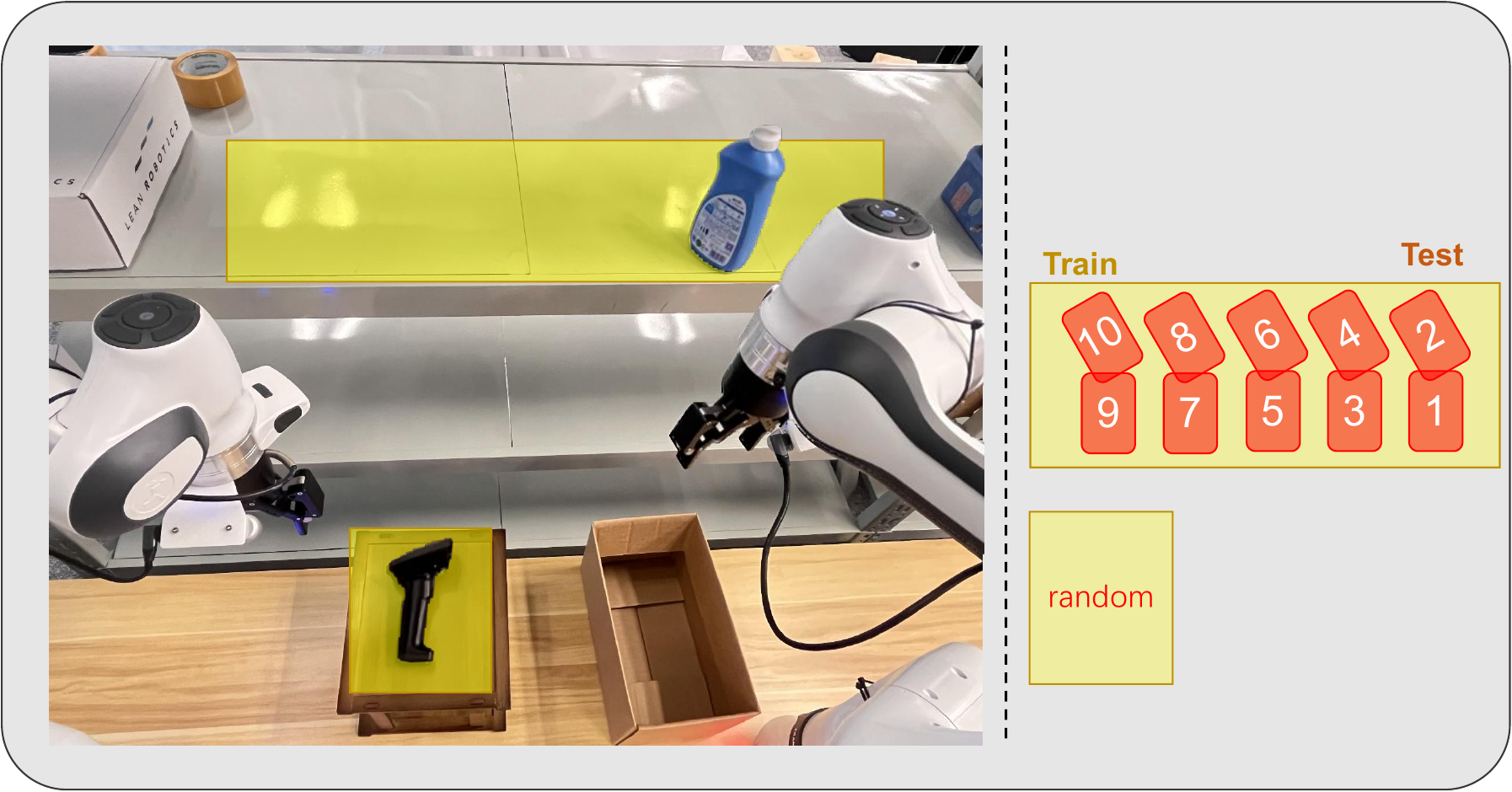}
    \caption{\textbf{Detailed 10 settings of Scan the Bottle.}
    Among the 10 settings, the bottle changes its position within the 10 red areas on the shelf, with variations in both position and rotation, while the scanner is placed randomly in minor adjustments on the platform.}
    \label{fig:exp-scan}
\end{figure}

As shown in Figure~\ref{fig:exp-carry}-\ref{fig:exp-scan}, the yellow area represents the variations during training, and the red area represents the settings during testing.
We set 10 settings for each task and every method has 3 trials for each setting.

\begin{table*}[htbp]
\centering
\caption{\textbf{Generalization tests across all settings.}}
\label{rebuttal-gen}
\resizebox{\linewidth}{!}{

\begin{tabular}{l*{9}{c}}
\toprule
\multirow{2}{*}{Method} 
& \multicolumn{3}{c}{Unseen Objects} 
& \multicolumn{3}{c}{Lighting Backgrounds} 
& \multicolumn{3}{c}{Object Interference} \\
\cmidrule(lr){2-4} \cmidrule(lr){5-7} \cmidrule(lr){8-10}
& Carry & Handover \& Insert & Wipe 
& Carry & Handover \& Insert & Wipe 
& Carry & Handover \& Insert & Wipe \\
\midrule
ACT     
& \redxmark / \redxmark 
& \redxmark / \redxmark 
& \redxmark / \redxmark 
& \redxmark / \redxmark 
& \redxmark / \redxmark 
& \redxmark/  \redxmark
& \greencheckmark / \greencheckmark 
& \redxmark / \redxmark 
& \redxmark / \redxmark 
\\
DP3     
& \greencheckmark / \redxmark
& \redxmark / \greencheckmark 
& \greencheckmark / \redxmark 
& \greencheckmark / \greencheckmark
& \redxmark / \greencheckmark
& \greencheckmark / \greencheckmark
& \redxmark / \redxmark 
& \redxmark / \redxmark
& \redxmark / \redxmark
\\
3D Diffuser Actor   
& \redxmark / \redxmark
& \redxmark / \redxmark
& \redxmark / \redxmark
& \redxmark / \redxmark
& \redxmark / \redxmark
& \redxmark / \redxmark
& \redxmark / \redxmark
& \redxmark / \redxmark
& \redxmark / \redxmark
\\
\baseline{Ours}
& \baseline{\greencheckmark / \greencheckmark}
& \baseline{\redxmark / \greencheckmark}
& \baseline{\greencheckmark / \redxmark}
& \baseline{\greencheckmark / \redxmark}
& \baseline{\greencheckmark / \greencheckmark}
& \baseline{\greencheckmark / \greencheckmark}
& \baseline{\greencheckmark / \greencheckmark}
& \baseline{\greencheckmark / \redxmark}
& \baseline{\greencheckmark / \redxmark}
\\
\bottomrule
\end{tabular}
}
\end{table*}


\subsection{Detailed Real-world Results}
We define three metrics: Success Rate (SR), Localization Success Rate (Loc-SR), and Normalized Score. 
And each task has several intermediate stages.

Specifically, \textbf{Carry the Tray} consists of four steps: (1) Pick the Tray, (2) Lift the Tray, (3) Place the Tray, (4) Return.
The Loc\_SR is recorded as 100\% only if the first two steps are achieved.

\textbf{Handover and Insert the Plate} involves three steps: (1) Pick the Plate, (2) Handover, (3) Insert the Plate.
The Loc\_SR is marked as 100\% only if the robot successfully picks up the plate.

\textbf{Wipe the Plate} consists of the following four steps: (1) Pick the Sponge, (2) Pick the Plate, (3) Wipe, (4) Place.
The Loc\_SR is recorded as 100\% only if both the sponge and the plate are successfully picked up.

For \textbf{Scan the Bottle}, the sequence is as follows: (1) Pick the Bottle, (2) Pick the Scanner, (3) Scan, (4) Place Bottle in Box.
Likewise, the Loc-SR is marked as 100\% only when both the bottle and the scanner are successfully picked up.

In \textbf{Wear the Scarf}, the arms need to carry out the following steps: (1) Pick the Scarf, (2) Lift the Scarf, (3) Wear the Scarf. 
If the scarf is picked up successfully, the Loc\_SR will be recorded as 100\%.

\textbf{Press the Cap of Bottle} also includes three motions: (1) Pick the Bottle, (2) Place the Bottle, (3) Press the Cap.
The Loc\_SR will be 100\% if the robot successfully picks up the bottle and places it on the cube.

For the single-arm experiment \textbf{Scan the Bottle with Single Arm}, the procedure comprises three stages: (1) Pick the Scanner, (2) Scan, (3) Place the Scanner.
The Loc\_SR is marked as 100\% if the first stage is completed.

Additionally, the score increases by 1 point upon the completion of each stage. 
To facilitate comparison, the total score is then normalized to a scale of 10 points, resulting in the Normalized Score.
The detailed results from the real-world experiments are available in Table~\ref{app_tray}-\ref{app_single}

\subsection{Detailed Real-world Generalization Results}

In the real-world generalization experiments, there are two settings for each generalization scenario (unseen objects, lighting background changes and object interference), with each setting undergoing 3 trials. 
The scoring criteria (SR, Loc\_SR and Score) and detailed real-world results remain consistent across these trials.

Additionally, we add all generalization tests to the first three tasks, all outperforming the baselines.
There are two different settings in each of these generalization tests. \textbf{Unseen objects}: Vary tray/plate shapes/colors (\textit{Carry}/\textit{Handover}); replace sponge with rag and round plate with rectangular one (\textit{Wipe}). \textbf{Lighting}: Dark and flickering lighting environments. \textbf{Object interference}: Introduce different distractor objects.
Results are summarized in Table~\ref{rebuttal-gen}.


The detailed results from the real-world generalization experiments are available in Table~\ref{app_gen_tray}-\ref{app_gen_scan}.

\begin{table*}[htbp]
\centering
\caption{\textbf{Detailed results on Carry the Tray task.}
}
\label{app_tray}
\scalebox{0.8}{\begin{tabular}{ccccccc}
\toprule

Method & Case Index  & Pick the Tray  & Lift the Tray &  Place the Tray  & Return & SR (\%) $\uparrow$ /Loc-SR (\%) $\uparrow$ / Score $\uparrow$ \\
\midrule \midrule 
\multirow{10}{*}{PPI}&1 
& \greencheckmark
& \greencheckmark
& \greencheckmark
& \redxmark
& 0.0 / 1.0 / 3
  \\
&2
& \greencheckmark
& \greencheckmark
& \greencheckmark
& \greencheckmark
& 1.0 / 1.0 / 4
 \\
&3
& \greencheckmark
& \greencheckmark
& \redxmark
& \redxmark
& 0.0 / 1.0 / 2
 \\
&4
& \greencheckmark
& \greencheckmark
& \greencheckmark
& \greencheckmark
& 1.0 / 1.0 / 4
  \\
&5
& \greencheckmark
& \greencheckmark
& \redxmark
& \redxmark
& 0.0 / 1.0 / 2 
  \\
&6
& \greencheckmark
& \greencheckmark
& \greencheckmark
& \greencheckmark
& 1.0 / 1.0 / 4 
 \\
&7
& \greencheckmark
& \greencheckmark
& \redxmark
& \redxmark
& 0.0 / 1.0 / 2
  \\
&8
& \greencheckmark
& \greencheckmark
& \greencheckmark
& \greencheckmark
& 1.0 / 1.0 / 4 
  \\
&9
& \greencheckmark
& \greencheckmark
& \greencheckmark
& \greencheckmark
& 1.0 / 1.0 / 4
  \\
&10
& \greencheckmark
& \greencheckmark
& \redxmark
&\redxmark
& 0.0 / 1.0 / 2 
 \\
\midrule \midrule
\multirow{10}{*}{Simple PPI}
&1 
& \greencheckmark
& \greencheckmark
& \greencheckmark
& \greencheckmark
& 1.0 / 1.0 / 4
  \\
&2
& \greencheckmark
& \greencheckmark
& \greencheckmark
& \greencheckmark
& 1.0 / 1.0 / 4
 \\
&3
& \greencheckmark
& \greencheckmark
& \greencheckmark
& \greencheckmark
& 1.0 / 1.0 / 4
 \\
&4
& \redxmark
& \redxmark
& \redxmark
& \redxmark
& 0.0 / 0.0 / 0
  \\
&5
& \redxmark
& \redxmark
& \redxmark
& \redxmark
& 0.0 / 0.0 / 0
  \\
&6
& \greencheckmark
& \greencheckmark
& \redxmark
& \redxmark
& 0.0 / 1.0 / 2 
 \\
&7
& \greencheckmark
& \greencheckmark
& \redxmark
& \redxmark
& 0.0 / 1.0 / 2
  \\
&8
& \greencheckmark
& \greencheckmark
& \greencheckmark
& \greencheckmark
& 1.0 / 1.0 / 4 
  \\
&9
& \greencheckmark
& \greencheckmark
& \greencheckmark
& \greencheckmark
& 1.0 / 1.0 / 4
  \\
&10
& \redxmark
& \redxmark
& \redxmark
&\redxmark
& 0.0 / 0.0 / 0 
 \\
\midrule \midrule
\multirow{10}{*}{ACT}&1 
& \greencheckmark
& \redxmark
& \redxmark
& \redxmark
& 0.0 / 0.0 / 1
  \\
&2
& \greencheckmark
& \redxmark
&\redxmark
& \redxmark
& 0.0 / 0.0 / 1
 \\
&3
& \greencheckmark
& \redxmark
&\redxmark
& \redxmark
& 0.0 / 0.0 / 1
 \\
&4
& \greencheckmark
& \greencheckmark
& \greencheckmark
& \greencheckmark
& 1.0 / 1.0 / 4
  \\
&5
& \greencheckmark
& \redxmark
&\redxmark
& \redxmark
& 0.0 / 0.0 / 1
  \\
&6
& \greencheckmark
& \redxmark
&\redxmark
& \redxmark
& 0.0 / 0.0 / 1
 \\
&7
& \greencheckmark
& \greencheckmark
& \greencheckmark
& \greencheckmark
& 1.0 / 1.0 / 4
  \\
&8
& \greencheckmark
& \greencheckmark
& \greencheckmark
& \greencheckmark
& 1.0 / 1.0 / 4
  \\
&9
& \greencheckmark
& \greencheckmark
& \greencheckmark
& \greencheckmark
& 1.0 / 1.0 / 4
  \\
&10
& \greencheckmark
& \redxmark
& \redxmark
&\redxmark
& 0.0 / 0.0 / 1 
 \\
\midrule \midrule
\multirow{10}{*}{DP3}&1 
& \greencheckmark
& \redxmark
& \redxmark
& \redxmark
& 0.0 / 0.0 / 1
  \\
&2
& \greencheckmark
& \greencheckmark
& \greencheckmark
& \greencheckmark
& 1.0 / 1.0 / 4
 \\
&3
& \greencheckmark
& \greencheckmark
& \greencheckmark
& \greencheckmark
& 1.0 / 1.0 / 4
 \\
&4
& \greencheckmark
& \greencheckmark
& \greencheckmark
& \greencheckmark
& 1.0 / 1.0 / 4
  \\
&5
& \greencheckmark
& \greencheckmark
& \greencheckmark
& \greencheckmark
& 1.0 / 1.0 / 4
  \\
&6
& \greencheckmark
& \redxmark
&\redxmark
& \redxmark
& 0.0 / 0.0 / 1
 \\
&7
& \greencheckmark
& \redxmark
&\redxmark
& \redxmark
& 0.0 / 0.0 / 1
  \\
&8
& \greencheckmark
& \greencheckmark
& \greencheckmark
& \greencheckmark
& 1.0 / 1.0 / 4
  \\
&9
& \greencheckmark
& \redxmark
&\redxmark
& \redxmark
& 0.0 / 0.0 / 1
  \\
&10
& \redxmark
& \redxmark
& \redxmark
& \redxmark
& 0.0 / 0.0 / 0
 \\
\midrule \midrule
\multirow{10}{*}{3D Diffuser Actor}&1 
& \greencheckmark
& \greencheckmark
& \redxmark
& \redxmark
& 0.0 / 1.0 / 2
  \\
&2
& \greencheckmark
& \greencheckmark
& \redxmark
& \redxmark
& 0.0 / 1.0 / 2
 \\
&3
& \greencheckmark
& \greencheckmark
& \redxmark
& \redxmark
& 0.0 / 1.0 / 2
 \\
&4
& \greencheckmark
& \greencheckmark
& \redxmark
& \redxmark
& 0.0 / 1.0 / 2
  \\
&5
& \greencheckmark
& \greencheckmark
& \greencheckmark
& \redxmark
& 0.0 / 1.0 / 3
  \\
&6
& \greencheckmark
& \redxmark
&\redxmark
& \redxmark
& 0.0 / 0.0 / 1
 \\
&7
& \greencheckmark
& \redxmark
&\redxmark
& \redxmark
& 0.0 / 0.0 / 1
  \\
&8
& \greencheckmark
& \redxmark
&\redxmark
& \redxmark
& 0.0 / 0.0 / 1
  \\
&9
& \greencheckmark
& \greencheckmark
&\redxmark
& \redxmark
& 0.0 / 1.0 / 2
  \\
&10
& \greencheckmark
& \greencheckmark
&\redxmark
& \redxmark
& 0.0 / 1.0 / 2
 \\
\bottomrule
\end{tabular}}
\vspace{-5mm}
\end{table*}

\begin{table*}[htbp]
\centering
\caption{\textbf{Detailed results on Handover and Insert the Plate task.}
}
\label{app_handover}
\scalebox{0.8}{\begin{tabular}{cccccc}
\toprule

Method & Case Index  & Pick the Plate & Handover &  Insert the Plate  & SR (\%) $\uparrow$ /Loc-SR (\%) $\uparrow$ / Score $\uparrow$ \\
\midrule \midrule 
\multirow{10}{*}{PPI}&1 
& \greencheckmark
& \greencheckmark
& \redxmark
& 0.0 / 1.0 / 2
  \\
&2
& \greencheckmark
& \greencheckmark
& \greencheckmark
& 1.0 / 1.0 / 3
 \\
&3
& \greencheckmark
& \greencheckmark
& \redxmark
& 0.0 / 1.0 / 2
 \\
&4
& \greencheckmark
& \greencheckmark
& \redxmark
& 0.0 / 1.0 / 2
  \\
&5
& \greencheckmark
& \greencheckmark
& \redxmark
& 0.0 / 1.0 / 2
  \\
&6
& \greencheckmark
& \redxmark
& \redxmark
& 0.0 / 1.0 / 1
 \\
&7
& \greencheckmark
& \greencheckmark
& \greencheckmark
& 1.0 / 1.0 / 3
  \\
&8
& \greencheckmark
& \greencheckmark
& \greencheckmark
& 1.0 / 1.0 / 3
  \\
&9
& \greencheckmark
& \greencheckmark
& \redxmark
& 0.0 / 1.0 / 2
  \\
&10
& \greencheckmark
& \greencheckmark
& \greencheckmark
& 1.0 / 1.0 / 3
 \\
\midrule \midrule
\multirow{10}{*}{Simple PPI}&1 
& \greencheckmark
& \redxmark
& \redxmark
& 0.0 / 1.0 / 1
  \\
&2
& \greencheckmark
& \greencheckmark
& \greencheckmark
& 1.0 / 1.0 / 3
 \\
&3
& \greencheckmark
& \redxmark
& \redxmark
& 0.0 / 1.0 / 1
 \\
&4
& \greencheckmark
& \redxmark
& \redxmark
& 0.0 / 1.0 / 1
  \\
&5
& \greencheckmark
& \greencheckmark
& \redxmark
& 0.0 / 1.0 / 2
  \\
&6
& \greencheckmark
& \redxmark
& \redxmark
& 0.0 / 1.0 / 1
 \\
&7
& \greencheckmark
& \greencheckmark
& \greencheckmark
& 1.0 / 1.0 / 3
  \\
&8
& \greencheckmark
& \greencheckmark
& \greencheckmark
& 1.0 / 1.0 / 3
  \\
&9
& \greencheckmark
& \redxmark
& \redxmark
& 0.0 / 1.0 / 1
  \\
&10
& \greencheckmark
& \redxmark
& \redxmark
& 0.0 / 1.0 / 1
 \\
\midrule \midrule
\multirow{10}{*}{ACT}&1 
& \redxmark
&\redxmark
& \redxmark
& 0.0 / 0.0 / 0
  \\
&2
& \redxmark
&\redxmark
& \redxmark
& 0.0 / 0.0 / 0
 \\
&3
& \greencheckmark
& \greencheckmark
& \redxmark
& 0.0 / 1.0 / 2
 \\
&4
& \greencheckmark
& \greencheckmark
& \greencheckmark
& 1.0 / 1.0 / 3
  \\
&5
& \greencheckmark
& \greencheckmark
& \redxmark
& 0.0 / 1.0 / 2
  \\
&6
& \greencheckmark
& \greencheckmark
& \redxmark
& 0.0 / 1.0 / 2
 \\
&7
& \greencheckmark
& \greencheckmark
& \redxmark
& 0.0 / 1.0 / 2
  \\
&8
& \greencheckmark
& \greencheckmark
& \redxmark
& 0.0 / 1.0 / 2
  \\
&9
& \greencheckmark
& \greencheckmark
& \redxmark
& 0.0 / 1.0 / 2
  \\
&10
& \redxmark
& \redxmark
& \redxmark
& 0.0 / 0.0 / 0
 \\
\midrule \midrule
\multirow{10}{*}{DP3}&1 
& \greencheckmark
& \redxmark
& \redxmark
& 0.0 / 1.0 / 1
  \\
&2
& \greencheckmark
& \greencheckmark
& \redxmark
& 0.0 / 1.0 / 2
 \\
&3
& \greencheckmark
& \greencheckmark
& \greencheckmark
& 1.0 / 1.0 / 3
 \\
&4
& \greencheckmark
& \redxmark
& \redxmark
& 0.0 / 1.0 / 1
  \\
&5
& \greencheckmark
& \greencheckmark
& \greencheckmark
& 1.0 / 1.0 / 3
  \\
&6
& \greencheckmark
&\redxmark
& \redxmark
& 0.0 / 1.0 / 1
 \\
&7
& \greencheckmark
& \greencheckmark
& \redxmark
& 0.0 / 1.0 / 2
  \\
&8
& \greencheckmark
&\redxmark
& \redxmark
& 0.0 / 1.0 / 1
  \\
&9
& \redxmark
& \redxmark
& \redxmark
& 0.0 / 0.0 / 0
  \\
&10
& \redxmark
& \redxmark
& \redxmark
& 0.0 / 0.0 / 0
 \\
\midrule \midrule
\multirow{10}{*}{3D Diffuser Actor}&1 
& \greencheckmark
& \redxmark
& \redxmark
& 0.0 / 1.0 / 1
  \\
&2
& \greencheckmark
& \greencheckmark
& \redxmark
& 0.0 / 1.0 / 2
 \\
&3
& \greencheckmark
& \redxmark
& \redxmark
& 0.0 / 1.0 / 1
 \\
&4
& \greencheckmark
& \greencheckmark
& \redxmark
& 0.0 / 1.0 / 2
  \\
&5
& \greencheckmark
& \greencheckmark
& \redxmark
& 0.0 / 1.0 / 2
  \\
&6
& \greencheckmark
& \greencheckmark
& \redxmark
& 0.0 / 1.0 / 2
 \\
&7
& \greencheckmark
& \greencheckmark
& \redxmark
& 0.0 / 1.0 / 2
  \\
&8
& \greencheckmark
& \greencheckmark
& \redxmark
& 0.0 / 1.0 / 2
  \\
&9
& \greencheckmark
& \greencheckmark
& \redxmark
& 0.0 / 1.0 / 2
  \\
&10
& \greencheckmark
& \greencheckmark
& \redxmark
& 0.0 / 1.0 / 2
 \\
\bottomrule
\end{tabular}}
\vspace{-5mm}
\end{table*}
\begin{table*}[htbp]
\centering
\caption{\textbf{Detailed results on Wipe the Plate task.}
}
\label{app_wipe}
\scalebox{0.8}{\begin{tabular}{ccccccc}
\toprule

Method & Case Index  & Pick the Sponge  & Pick the Plate &  Wipe  & Place & SR (\%) $\uparrow$ /Loc-SR (\%) $\uparrow$ / Score $\uparrow$ \\
\midrule \midrule 
\multirow{10}{*}{PPI}&1 
& \greencheckmark
& \greencheckmark
& \greencheckmark
& \redxmark
& 0.0 / 1.0 / 3
  \\
&2
& \greencheckmark
& \greencheckmark
& \greencheckmark
& \greencheckmark
& 1.0 / 1.0 / 4
 \\
&3
& \greencheckmark
& \redxmark
& \redxmark
& \redxmark
& 0.0 / 0.0 / 1
 \\
&4
& \greencheckmark
& \greencheckmark
& \greencheckmark
& \greencheckmark
& 1.0 / 1.0 / 4
  \\
&5
& \greencheckmark
& \greencheckmark
& \greencheckmark
& \greencheckmark
& 1.0 / 1.0 / 4
  \\
&6
& \greencheckmark
& \greencheckmark
& \greencheckmark
& \greencheckmark
& 1.0 / 1.0 / 4 
 \\
&7
& \greencheckmark
& \redxmark
& \redxmark
& \redxmark
& 0.0 / 0.0 / 1
  \\
&8
& \greencheckmark
& \greencheckmark
& \greencheckmark
& \greencheckmark
& 1.0 / 1.0 / 4 
  \\
&9
& \greencheckmark
& \greencheckmark
& \greencheckmark
& \greencheckmark
& 1.0 / 1.0 / 4
  \\
&10
& \greencheckmark
& \greencheckmark
& \greencheckmark
& \greencheckmark
& 1.0 / 1.0 / 4 
 \\
\midrule \midrule
\multirow{10}{*}{Simple PPI}&1 
& \greencheckmark
& \greencheckmark
& \greencheckmark
& \greencheckmark
& 1.0 / 1.0 / 4
  \\
&2
& \greencheckmark
& \redxmark
& \redxmark
& \redxmark
& 0.0 / 0.0 / 1
 \\
&3
& \greencheckmark
& \greencheckmark
& \redxmark
& \redxmark
& 0.0 / 1.0 / 2
 \\
&4
& \greencheckmark
& \greencheckmark
& \greencheckmark
& \greencheckmark
& 1.0 / 1.0 / 4
  \\
&5
& \greencheckmark
& \greencheckmark
& \greencheckmark
& \greencheckmark
& 1.0 / 1.0 / 4
  \\
&6
& \greencheckmark
& \greencheckmark
& \greencheckmark
& \greencheckmark
& 1.0 / 1.0 / 4 
 \\
&7
& \greencheckmark
& \redxmark
& \redxmark
& \redxmark
& 0.0 / 0.0 / 1
  \\
&8
& \greencheckmark
& \redxmark
& \redxmark
& \redxmark
& 0.0 / 0.0 / 1
  \\
&9
& \greencheckmark
& \greencheckmark
& \redxmark
& \redxmark
& 0.0 / 1.0 / 2
  \\
&10
& \greencheckmark
& \greencheckmark
& \greencheckmark
& \greencheckmark
& 1.0 / 1.0 / 4 
 \\
\midrule \midrule
\multirow{10}{*}{ACT}&1 
& \greencheckmark
& \greencheckmark
& \redxmark
& \greencheckmark
& 0.0 / 1.0 / 3
  \\
&2
& \redxmark
& \greencheckmark
& \redxmark
& \redxmark
& 0.0 / 0.0 / 1
 \\
&3
& \greencheckmark
& \redxmark
& \redxmark
& \redxmark
& 0.0 / 0.0 / 1
 \\
&4
& \greencheckmark
& \greencheckmark
& \redxmark
& \redxmark
& 0.0 / 1.0 / 2
  \\
&5
& \greencheckmark
& \greencheckmark
& \redxmark
& \redxmark
& 0.0 / 1.0 / 2
  \\
&6
& \greencheckmark
& \redxmark
& \redxmark
& \redxmark
& 0.0 / 0.0 / 1
 \\
&7
& \greencheckmark
& \redxmark
& \redxmark
& \redxmark
& 0.0 / 0.0 / 1
  \\
&8
& \greencheckmark
& \redxmark
& \redxmark
& \redxmark
& 0.0 / 0.0 / 1
  \\
&9
& \greencheckmark
& \greencheckmark
& \redxmark
& \redxmark
& 0.0 / 1.0 / 2
  \\
&10
& \greencheckmark
& \redxmark
& \redxmark
& \redxmark
& 0.0 / 0.0 / 1
 \\
\midrule \midrule
\multirow{10}{*}{DP3}&1 
& \greencheckmark
& \greencheckmark
& \greencheckmark
& \greencheckmark
& 1.0 / 1.0 / 4
  \\
&2
& \greencheckmark
& \greencheckmark
& \greencheckmark
& \greencheckmark
& 1.0 / 1.0 / 4
 \\
&3
& \greencheckmark
& \greencheckmark
& \greencheckmark
& \greencheckmark
& 1.0 / 1.0 / 4
 \\
&4
& \greencheckmark
& \greencheckmark
& \greencheckmark
& \greencheckmark
& 1.0 / 1.0 / 4
  \\
&5
& \greencheckmark
& \greencheckmark
& \greencheckmark
& \redxmark
& 0.0 / 1.0 / 3
  \\
&6
& \redxmark
& \greencheckmark
& \redxmark
& \redxmark
& 0.0 / 0.0 / 1
 \\
&7
& \redxmark
& \greencheckmark
& \redxmark
& \redxmark
& 0.0 / 0.0 / 1
  \\
&8
& \redxmark
& \greencheckmark
& \redxmark
& \redxmark
& 0.0 / 0.0 / 1
  \\
&9
& \greencheckmark
& \greencheckmark
& \greencheckmark
& \redxmark
& 0.0 / 1.0 / 3
  \\
&10
& \redxmark
& \greencheckmark
& \redxmark
& \redxmark
& 0.0 / 0.0 / 1
 \\
\midrule \midrule
\multirow{10}{*}{3D Diffuser Actor}&1 
& \greencheckmark
& \redxmark
& \redxmark
& \redxmark
& 0.0 / 0.0 / 1
  \\
&2
& \greencheckmark
& \greencheckmark
& \redxmark
& \redxmark
& 0.0 / 1.0 / 2
 \\
&3
& \greencheckmark
& \redxmark
& \redxmark
& \redxmark
& 0.0 / 0.0 / 1
  \\
&4
& \greencheckmark
& \greencheckmark
& \redxmark
& \redxmark
& 0.0 / 1.0 / 2
  \\
&5
& \greencheckmark
& \redxmark
& \redxmark
& \redxmark
& 0.0 / 0.0 / 1
  \\
&6
& \greencheckmark
& \greencheckmark
& \redxmark
& \redxmark
& 0.0 / 1.0 / 2
 \\
&7
& \redxmark
& \redxmark
&\redxmark
& \redxmark
& 0.0 / 0.0 / 0
  \\
&8
& \greencheckmark
& \greencheckmark
&\redxmark
& \redxmark
& 0.0 / 1.0 / 2
  \\
&9
& \greencheckmark
& \redxmark
&\redxmark
& \redxmark
& 0.0 / 0.0 / 1
  \\
&10
& \greencheckmark
& \greencheckmark
&\redxmark
& \redxmark
& 0.0 / 1.0 / 2
 \\
\bottomrule
\end{tabular}}
\vspace{-5mm}
\end{table*}
\begin{table*}[htbp]
\centering
\caption{\textbf{Detailed results on Scan the Bottle task.}
}
\label{app_scan}
\scalebox{0.8}{\begin{tabular}{ccccccc}
\toprule

Method & Case Index  & Pick the Bottle  & Pick the Scanner &  Scan  & Place Bottle in Box & SR (\%) $\uparrow$ /Loc-SR (\%) $\uparrow$ / Score $\uparrow$ \\
\midrule \midrule 
\multirow{10}{*}{PPI}&1 
& \greencheckmark
& \greencheckmark
& \greencheckmark
& \greencheckmark
& 1.0 / 1.0 / 4
  \\
&2
& \greencheckmark
& \greencheckmark
& \greencheckmark
& \greencheckmark
& 1.0 / 1.0 / 4
 \\
&3
& \greencheckmark
& \greencheckmark
& \greencheckmark
& \greencheckmark
& 1.0 / 1.0 / 4
 \\
&4
& \greencheckmark
& \greencheckmark
& \greencheckmark
& \greencheckmark
& 1.0 / 1.0 / 4
  \\
&5
& \greencheckmark
& \greencheckmark
& \greencheckmark
& \greencheckmark
& 1.0 / 1.0 / 4
  \\
&6
& \greencheckmark
& \greencheckmark
& \greencheckmark
& \greencheckmark
& 1.0 / 1.0 / 4
 \\
&7
& \redxmark
& \greencheckmark
& \redxmark
& \redxmark
& 0.0 / 0.0 / 1
  \\
&8
& \greencheckmark
& \greencheckmark
& \greencheckmark
& \greencheckmark
& 1.0 / 1.0 / 4
  \\
&9
& \greencheckmark
& \greencheckmark
& \greencheckmark
& \greencheckmark
& 1.0 / 1.0 / 4
  \\
&10
& \greencheckmark
& \greencheckmark
& \greencheckmark
& \greencheckmark
& 1.0 / 1.0 / 4 
 \\
\midrule \midrule
\multirow{10}{*}{ACT}&1 
& \greencheckmark
& \greencheckmark
& \redxmark
& \redxmark
& 0.0 / 1.0 / 2
  \\
&2
& \redxmark
& \greencheckmark
& \redxmark
& \redxmark
& 0.0 / 0.0 / 1
 \\
&3
& \greencheckmark
& \greencheckmark
& \redxmark
& \redxmark
& 0.0 / 1.0 / 2
 \\
&4
& \greencheckmark
& \greencheckmark
& \greencheckmark
& \greencheckmark
& 1.0 / 1.0 / 4
  \\
&5
& \redxmark
& \greencheckmark
& \redxmark
& \redxmark
& 0.0 / 0.0 / 1
  \\
&6
& \greencheckmark
& \greencheckmark
& \redxmark
& \greencheckmark
& 0.0 / 1.0 / 3
 \\
&7
& \redxmark
& \greencheckmark
& \redxmark
& \redxmark
& 0.0 / 0.0 / 1

  \\
&8
& \greencheckmark
& \greencheckmark
& \redxmark
& \redxmark
& 0.0 / 1.0 / 2
  \\
&9
& \redxmark
& \redxmark
& \redxmark
& \redxmark
& 0.0 / 0.0 / 0
  \\
&10
& \redxmark
& \greencheckmark
& \redxmark
& \redxmark
& 0.0 / 0.0 / 1
 \\
\midrule \midrule
\multirow{10}{*}{DP3}&1 
& \greencheckmark
& \redxmark
& \redxmark
& \redxmark
& 0.0 / 0.0 / 1
  \\
&2
& \greencheckmark
& \greencheckmark
& \greencheckmark
& \greencheckmark
& 1.0 / 1.0 / 4
 \\
&3
& \greencheckmark
& \greencheckmark
& \redxmark
& \redxmark
& 0.0 / 1.0 / 2
 \\
&4
& \redxmark
& \greencheckmark
& \redxmark
& \redxmark
& 0.0 / 0.0 / 1
  \\
&5
& \redxmark
& \greencheckmark
& \redxmark
& \redxmark
& 0.0 / 0.0 / 1
  \\
&6
& \redxmark
& \greencheckmark
& \redxmark
& \redxmark
& 0.0 / 0.0 / 1
 \\
&7
& \greencheckmark
& \greencheckmark
&\greencheckmark
& \greencheckmark
& 1.0 / 1.0 / 4
  \\
&8
& \redxmark
& \redxmark
&\redxmark
& \redxmark
& 0.0 / 0.0 / 0
  \\
&9
& \greencheckmark
& \redxmark
&\redxmark
& \redxmark
& 0.0 / 0.0 / 1
  \\
&10
& \greencheckmark
& \greencheckmark
&\greencheckmark
& \greencheckmark
& 1.0 / 1.0 / 4
 \\
\midrule \midrule
\multirow{10}{*}{3D Diffuser Actor}&1 
& \redxmark
& \greencheckmark
&\redxmark
& \redxmark
& 0.0 / 0.0 / 1
  \\
&2
& \greencheckmark
& \redxmark
& \redxmark
& \redxmark
& 0.0 / 0.0 / 1
 \\
&3
& \greencheckmark
& \greencheckmark
& \greencheckmark
& \greencheckmark
& 1.0 / 1.0 / 4
  \\
&4
& \greencheckmark
& \greencheckmark
& \redxmark
& \greencheckmark
& 0.0 / 1.0 / 3
  \\
&5
& \redxmark
& \greencheckmark
&\redxmark
& \redxmark
& 0.0 / 0.0 / 1
  \\
&6
& \greencheckmark
& \redxmark
&\redxmark
& \redxmark
& 0.0 / 0.0 / 1
 \\
&7
& \greencheckmark
& \greencheckmark
& \greencheckmark
& \greencheckmark
& 1.0 / 1.0 / 4
  \\
&8
& \greencheckmark
& \greencheckmark
&\redxmark
& \redxmark
& 0.0 / 1.0 / 2
  \\
&9
& \redxmark
& \greencheckmark
&\redxmark
& \redxmark
& 0.0 / 0.0 / 1
  \\
&10
& \redxmark
& \greencheckmark
&\redxmark
& \redxmark
& 0.0 / 0.0 / 1
 \\
\bottomrule
\end{tabular}}
\vspace{-5mm}
\end{table*}

\begin{table*}[htbp]
\centering
\caption{\textbf{Detailed results on Wear the Scarf task.}
}
\label{app_wear}
\scalebox{0.6}{\begin{tabular}{cccccc}
\toprule

Method & Case Index  & Pick the Scarf & Lift the Scarf &  Wear the Scarf & SR (\%) $\uparrow$ /Loc-SR (\%) $\uparrow$ / Score $\uparrow$ \\
\midrule \midrule 
\multirow{10}{*}{PPI}&1 
& \greencheckmark
& \greencheckmark
& \greencheckmark
& 1.0 / 1.0 / 3
  \\
&2
& \greencheckmark
& \greencheckmark
& \greencheckmark
& 1.0 / 1.0 / 3
 \\
&3
& \greencheckmark
& \greencheckmark
& \redxmark
& 0.0 / 1.0 / 2
 \\
&4
& \redxmark
& \redxmark
& \redxmark
& 0.0 / 0.0 / 0
  \\
&5
& \greencheckmark
& \greencheckmark
& \greencheckmark
& 1.0 / 1.0 / 3
  \\
&6
& \greencheckmark
& \greencheckmark
& \greencheckmark
& 1.0 / 1.0 / 3
 \\
&7
& \greencheckmark
& \greencheckmark
& \greencheckmark
& 1.0 / 1.0 / 3
  \\
&8
& \greencheckmark
& \greencheckmark
& \greencheckmark
& 1.0 / 1.0 / 3
  \\
&9
& \redxmark
& \redxmark
& \redxmark
& 0.0 / 0.0 / 0
  \\
&10
& \greencheckmark
& \greencheckmark
& \greencheckmark
& 1.0 / 1.0 / 3
 \\
\midrule \midrule
\multirow{10}{*}{ACT}&1 
& \redxmark
&\redxmark
& \redxmark
& 0.0 / 0.0 / 0
  \\
&2
& \greencheckmark
& \greencheckmark
& \greencheckmark
& 1.0 / 1.0 / 3
 \\
&3
& \greencheckmark
& \greencheckmark
& \greencheckmark
& 1.0 / 1.0 / 3
 \\
&4
& \redxmark
&\redxmark
& \redxmark
& 0.0 / 0.0 / 0
  \\
&5
& \greencheckmark
& \greencheckmark
& \greencheckmark
& 1.0 / 1.0 / 3
  \\
&6
& \redxmark
&\redxmark
& \redxmark
& 0.0 / 0.0 / 0
 \\
&7
& \greencheckmark
& \greencheckmark
& \redxmark
& 0.0 / 1.0 / 2
  \\
&8
& \greencheckmark
& \greencheckmark
& \greencheckmark
& 1.0 / 1.0 / 3
  \\
&9
& \greencheckmark
& \redxmark
& \redxmark
& 0.0 / 1.0 / 1
  \\
&10
& \greencheckmark
& \greencheckmark
& \redxmark
& 0.0 / 1.0 / 2
 \\
\bottomrule
\end{tabular}}
\vspace{-5mm}
\end{table*}

\begin{table*}[htbp]
\centering
\caption{\textbf{Detailed results on Press the Cap of Bottle task.}
}
\label{app_press}
\scalebox{0.6}{\begin{tabular}{cccccc}
\toprule

Method & Case Index  & Pick the Bottle & Place the Bottle &  Press the Cap & SR (\%) $\uparrow$ /Loc-SR (\%) $\uparrow$ / Score $\uparrow$ \\
\midrule \midrule 
\multirow{10}{*}{PPI}&1 
& \greencheckmark
& \greencheckmark
& \greencheckmark
& 1.0 / 1.0 / 3
  \\
&2
& \greencheckmark
& \greencheckmark
& \greencheckmark
& 1.0 / 1.0 / 3
 \\
&3
& \greencheckmark
& \greencheckmark
& \redxmark
& 0.0 / 1.0 / 2
 \\
&4
& \redxmark
&\redxmark
& \redxmark
& 0.0 / 0.0 / 0
  \\
&5
& \greencheckmark
& \greencheckmark
& \redxmark
& 0.0 / 1.0 / 2
  \\
&6
& \redxmark
&\redxmark
& \redxmark
& 0.0 / 0.0 / 0
 \\
&7
& \greencheckmark
& \greencheckmark
& \redxmark
& 0.0 / 1.0 / 2
  \\
&8
& \greencheckmark
& \greencheckmark
& \greencheckmark
& 1.0 / 1.0 / 3
  \\
&9
& \greencheckmark
& \greencheckmark
& \greencheckmark
& 1.0 / 1.0 / 3
  \\
&10
& \greencheckmark
& \greencheckmark
& \redxmark
& 0.0 / 1.0 / 2
 \\
\midrule \midrule
\multirow{10}{*}{ACT}&1 
& \greencheckmark
&\redxmark
& \redxmark
& 0.0 / 0.0 / 1
  \\
&2
& \greencheckmark
&\redxmark
& \redxmark
& 0.0 / 0.0 / 1
 \\
&3
& \greencheckmark
&\redxmark
& \redxmark
& 0.0 / 0.0 / 1
 \\
&4
& \greencheckmark
& \greencheckmark
& \greencheckmark
& 1.0 / 1.0 / 3
  \\
&5
& \greencheckmark
&\redxmark
& \redxmark
& 0.0 / 0.0 / 1
  \\
&6
& \redxmark
& \redxmark
& \redxmark
& 0.0 / 0.0 / 0
 \\
&7
& \greencheckmark
& \greencheckmark
& \redxmark
& 0.0 / 1.0 / 2
  \\
&8
& \redxmark
& \redxmark
& \redxmark
& 0.0 / 0.0 / 0
  \\
&9
& \redxmark
& \redxmark
& \redxmark
& 0.0 / 0.0 / 0
  \\
&10
& \redxmark
& \redxmark
& \redxmark
& 0.0 / 0.0 / 0
 \\
\bottomrule
\end{tabular}}
\vspace{-5mm}
\end{table*}

\begin{table*}[htbp]
\centering
\caption{\textbf{Detailed results on Scan the Bottle with Single Arm task.}
}
\label{app_single}
\scalebox{0.6}{\begin{tabular}{cccccc}
\toprule

Method & Case Index  & Pick the Scanner & Scan &  Place the Scanner & SR (\%) $\uparrow$ /Loc-SR (\%) $\uparrow$ / Score $\uparrow$ \\
\midrule \midrule 
\multirow{10}{*}{PPI}&1 
& \greencheckmark
& \greencheckmark
& \greencheckmark
& 1.0 / 1.0 / 3
  \\
&2
& \greencheckmark
& \greencheckmark
& \greencheckmark
& 1.0 / 1.0 / 3
 \\
&3
& \greencheckmark
& \greencheckmark
& \greencheckmark
& 1.0 / 1.0 / 3
 \\
&4
& \greencheckmark
& \greencheckmark
& \greencheckmark
& 1.0 / 1.0 / 3
  \\
&5
& \greencheckmark
& \redxmark
& \greencheckmark
& 0.0 / 1.0 / 2
  \\
&6
& \greencheckmark
& \greencheckmark
& \greencheckmark
& 1.0 / 1.0 / 3
 \\
&7
& \greencheckmark
& \redxmark
& \greencheckmark
& 0.0 / 1.0 / 2
  \\
&8
& \greencheckmark
& \greencheckmark
& \greencheckmark
& 1.0 / 1.0 / 3
  \\
&9
& \greencheckmark
& \greencheckmark
& \greencheckmark
& 1.0 / 1.0 / 3
  \\
&10
& \greencheckmark
& \greencheckmark
& \greencheckmark
& 1.0 / 1.0 / 3
 \\
\midrule \midrule
\multirow{10}{*}{ACT}&1 
& \greencheckmark
&\redxmark
& \redxmark
& 0.0 / 1.0 / 1
  \\
&2
& \greencheckmark
&\redxmark
& \redxmark
& 0.0 / 1.0 / 1
 \\
&3
& \redxmark
&\redxmark
& \redxmark
& 0.0 / 0.0 / 0
 \\
&4
& \redxmark
&\redxmark
& \redxmark
& 0.0 / 0.0 / 0
  \\
&5
& \greencheckmark
& \redxmark
& \greencheckmark
& 0.0 / 1.0 / 2
  \\
&6
& \greencheckmark
& \greencheckmark
& \greencheckmark
& 1.0 / 1.0 / 3
 \\
&7
& \redxmark
&\redxmark
& \redxmark
& 0.0 / 0.0 / 0
  \\
&8
& \greencheckmark
& \greencheckmark
& \greencheckmark
& 1.0 / 1.0 / 3
  \\
&9
& \greencheckmark
& \redxmark
& \greencheckmark
& 0.0 / 1.0 / 2
  \\
&10
& \greencheckmark
& \redxmark
& \greencheckmark
& 0.0 / 1.0 / 2
 \\
\bottomrule
\end{tabular}}
\vspace{-5mm}
\end{table*}

\begin{table*}[htbp]
\centering
\caption{\textbf{Detailed generalization experiment results on Carry the Tray task.}
}
\label{app_gen_tray}
\scalebox{0.6}{\begin{tabular}{cccccccc}
\toprule

Method & Generalization Type & Case Index  & Pick the Tray  & Lift the Tray &  Place the Tray  & Return & SR (\%) $\uparrow$ /Loc-SR (\%) $\uparrow$ / Score $\uparrow$ \\
\midrule \midrule 
\multirow{7}{*}{PPI}
& Normal
& Normal Setting
& \greencheckmark
& \greencheckmark
& \greencheckmark
& \greencheckmark
& 1.0 / 1.0 / 4
  \\
 & \multirow{2}{*}{Unseen Objects}
& Round Tray
& \greencheckmark
& \greencheckmark
& \greencheckmark
& \greencheckmark
& 1.0 / 1.0 / 4
 \\
 & & Brown Tray
& \greencheckmark
& \greencheckmark
& \greencheckmark
& \greencheckmark
& 1.0 / 1.0 / 4
  \\
 & \multirow{2}{*}{Lighting Changes}
& Dark
& \greencheckmark
& \greencheckmark
& \greencheckmark
& \greencheckmark
& 1.0 / 1.0 / 4
 \\
 & & Flickering
& \greencheckmark
& \greencheckmark
& \redxmark
& \redxmark
& 0.0 / 1.0 / 2
  \\
 & \multirow{2}{*}{Object Interference}
& Rubik’s Cubes 
& \greencheckmark
& \greencheckmark
& \greencheckmark
& \greencheckmark
& 1.0 / 1.0 / 4
 \\
 & & Colorful Cubes
& \greencheckmark
& \greencheckmark
& \greencheckmark
& \greencheckmark
& 1.0 / 1.0 / 4
 
 \\
\midrule \midrule
\multirow{7}{*}{ACT}
& Normal
& Normal Setting
& \greencheckmark
& \greencheckmark
& \greencheckmark
& \greencheckmark
& 1.0 / 1.0 / 4
  \\
 & \multirow{2}{*}{Unseen Objects}
& Round Tray
& \redxmark
& \redxmark
& \redxmark
& \redxmark
& 0.0 / 0.0 / 0
 \\
 & & Brown Tray
& \redxmark
& \redxmark
& \redxmark
& \redxmark
& 0.0 / 0.0 / 0
  \\
 & \multirow{2}{*}{Lighting Changes}
& Dark
& \redxmark
& \redxmark
& \redxmark
& \redxmark
& 0.0 / 0.0 / 0
 \\
 & & Flickering
& \redxmark
& \redxmark
& \redxmark
& \redxmark
& 0.0 / 0.0 / 0
  \\
 & \multirow{2}{*}{Object Interference}
& Rubik’s Cubes 
& \greencheckmark
& \greencheckmark
& \greencheckmark
& \greencheckmark
& 1.0 / 1.0 / 4
 \\
 & & Colorful Cubes
& \greencheckmark
& \greencheckmark
& \greencheckmark
& \greencheckmark
& 1.0 / 1.0 / 4
 
 \\
\midrule \midrule
\multirow{7}{*}{DP3}
& Normal
& Normal Setting
& \greencheckmark
& \greencheckmark
& \greencheckmark
& \greencheckmark
& 1.0 / 1.0 / 4
  \\
 & \multirow{2}{*}{Unseen Objects}
& Round Tray
& \greencheckmark
& \greencheckmark
& \greencheckmark
& \greencheckmark
& 1.0 / 1.0 / 4
 \\
 & & Brown Tray
& \greencheckmark
& \redxmark
& \redxmark
& \redxmark
& 0.0 / 0.0 / 1
  \\
 & \multirow{2}{*}{Lighting Changes}
& Dark
& \greencheckmark
& \greencheckmark
& \greencheckmark
& \greencheckmark
& 1.0 / 1.0 / 4
 \\
 & & Flickering
& \greencheckmark
& \greencheckmark
& \greencheckmark
& \greencheckmark
& 1.0 / 1.0 / 4
  \\
 & \multirow{2}{*}{Object Interference}
& Rubik’s Cubes 
& \greencheckmark
& \redxmark
& \redxmark
& \redxmark
& 0.0 / 0.0 / 1
 \\
 & & Colorful Cubes
& \greencheckmark
& \greencheckmark
& \redxmark
& \redxmark
& 0.0 / 1.0 / 2
 
 \\
\midrule \midrule
\multirow{7}{*}{3D Diffuser Actor}
& Normal
& Normal Setting
& \greencheckmark
& \greencheckmark
& \greencheckmark
& \redxmark
& 0.0 / 1.0 / 3
  \\
 & \multirow{2}{*}{Unseen Objects}
& Round Tray
& \greencheckmark
& \greencheckmark
& \redxmark
& \redxmark
& 0.0 / 1.0 / 2
 \\
 & & Brown Tray
& \greencheckmark
& \greencheckmark
& \redxmark
& \redxmark
& 0.0 / 1.0 / 2
  \\
 & \multirow{2}{*}{Lighting Changes}
& Dark
& \redxmark
& \redxmark
& \redxmark
& \redxmark
& 0.0 / 0.0 / 0
 \\
 & & Lighting
& \redxmark
& \redxmark
& \redxmark
& \redxmark
& 0.0 / 0.0 / 0
  \\
 & \multirow{2}{*}{Object Interference}
& Rubik’s Cubes 
& \greencheckmark
& \greencheckmark
& \redxmark
& \redxmark
& 0.0 / 1.0 / 2
 \\
 & & Colorful Cubes
& \greencheckmark
& \redxmark
& \redxmark
& \redxmark
& 0.0 / 0.0 / 1
 
 \\
\bottomrule
\end{tabular}}
\vspace{-5mm}
\end{table*}

\begin{table*}[htbp]
\centering
\caption{\textbf{Detailed generalization experiment results on Handover and Insert the Plate task.}
}
\label{app_gen_handover}
\scalebox{0.6}{\begin{tabular}{ccccccc}
\toprule

Method & Generalization Type & Case Index  & Pick the Plate  & Handover &  Insert the Plate & SR (\%) $\uparrow$ /Loc-SR (\%) $\uparrow$ / Score $\uparrow$ \\
\midrule \midrule 
\multirow{7}{*}{PPI}
& Normal
& Normal Setting
& \greencheckmark
& \greencheckmark
& \greencheckmark
& 1.0 / 1.0 / 3
  \\
 & \multirow{2}{*}{Unseen Objects}
& Rectangular Plate
& \greencheckmark
& \redxmark
& \redxmark
& 0.0 / 1.0 / 1
 \\
 & & Green Plate
& \greencheckmark
& \greencheckmark
& \greencheckmark
& 1.0 / 1.0 / 3
  \\
 & \multirow{2}{*}{Lighting Changes}
& Dark
& \greencheckmark
& \greencheckmark
& \greencheckmark
& 1.0 / 1.0 / 3
 \\
 & & Flickering
& \greencheckmark
& \greencheckmark
& \greencheckmark
& 1.0 / 1.0 / 3
  \\
 & \multirow{2}{*}{Object Interference}
& Colorful Cubes 
& \greencheckmark
& \greencheckmark
& \greencheckmark
& 1.0 / 1.0 / 3
 \\
 & & Multi-Plate
& \greencheckmark
& \greencheckmark
& \redxmark
& 0.0 / 1.0 / 2
 
 \\
\midrule \midrule
\multirow{7}{*}{ACT}
& Normal
& Normal Setting
& \greencheckmark
& \greencheckmark
& \greencheckmark
& 1.0 / 1.0 / 3
  \\
 & \multirow{2}{*}{Unseen Objects}
& Rectangular Plate
& \redxmark
& \redxmark
& \redxmark
& 0.0 / 0.0 / 0
 \\
 & & Green Plate
& \greencheckmark
& \redxmark
& \redxmark
& 0.0 / 1.0 / 1
  \\
 & \multirow{2}{*}{Lighting Changes}
& Dark
& \redxmark
& \redxmark
& \redxmark
& 0.0 / 0.0 / 0
 \\
 & & Flickering
& \redxmark
& \redxmark
& \redxmark
& 0.0 / 0.0 / 0
  \\
 & \multirow{2}{*}{Object Interference}
& Colorful Cubes 
& \greencheckmark
& \greencheckmark
& \redxmark
& 0.0 / 1.0 / 2
 \\
 & & Multi-Plate
& \greencheckmark
& \greencheckmark
& \redxmark
& 0.0 / 1.0 / 2
 
 \\
\midrule \midrule
\multirow{7}{*}{DP3}
& Normal
& Normal Setting
& \greencheckmark
& \greencheckmark
& \greencheckmark
& 1.0 / 1.0 / 3
  \\
 & \multirow{2}{*}{Unseen Objects}
& Rectangular Plate
& \greencheckmark
& \redxmark
& \redxmark
& 0.0 / 1.0 / 1
 \\
 & & Green Plate
& \greencheckmark
& \greencheckmark
& \greencheckmark
& 1.0 / 1.0 / 3
  \\
 & \multirow{2}{*}{Lighting Changes}
& Dark
& \greencheckmark
& \greencheckmark
& \redxmark
& 0.0 / 1.0 / 2
 \\
 & & Flickering
& \greencheckmark
& \greencheckmark
& \greencheckmark
& 1.0 / 1.0 / 3
  \\
 & \multirow{2}{*}{Object Interference}
& Colorful Cubes 
& \redxmark
& \redxmark
& \redxmark
& 0.0 / 0.0 / 0
 \\
 & & Multi-Plate
& \greencheckmark
& \redxmark
& \redxmark
& 0.0 / 1.0 / 1
 
 \\
\midrule \midrule
\multirow{7}{*}{3D Diffuser Actor}
& Normal
& Normal Setting
& \greencheckmark
& \greencheckmark
& \redxmark
& 0.0 / 1.0 / 2
  \\
 & \multirow{2}{*}{Unseen Objects}
& Rectangular Plate
& \greencheckmark
& \redxmark
& \redxmark
& 0.0 / 1.0 / 1
 \\
 & & Green Plate
& \greencheckmark
& \redxmark
& \redxmark
& 0.0 / 1.0 / 1
  \\
 & \multirow{2}{*}{Lighting Changes}
& Dark
& \redxmark
& \redxmark
& \redxmark
& 0.0 / 0.0 / 0
 \\
 & & Flickering
& \redxmark
& \redxmark
& \redxmark
& 0.0 / 0.0 / 0
  \\
 & \multirow{2}{*}{Object Interference}
& Colorful Cubes 
& \greencheckmark
& \redxmark
& \redxmark
& 0.0 / 1.0 / 1
 \\
 & & Multi-Plate
& \greencheckmark
& \redxmark
& \redxmark
& 0.0 / 1.0 / 1
 
 \\
\bottomrule
\end{tabular}}
\vspace{-5mm}
\end{table*}
\begin{table*}[htbp]
\centering
\caption{\textbf{Detailed generalization experiment results on Wipe the Plate task.}
}
\label{app_gen_wipe}
\scalebox{0.6}{\begin{tabular}{cccccccc}
\toprule

Method & Generalization Type & Case Index  & Pick the Sponge  & Pick the Plate &  Wipe  & Place & SR (\%) $\uparrow$ /Loc-SR (\%) $\uparrow$ / Score $\uparrow$ \\
\midrule \midrule 
\multirow{7}{*}{PPI}
& Normal
& Normal Setting
& \greencheckmark
& \greencheckmark
& \greencheckmark
& \greencheckmark
& 1.0 / 1.0 / 4
  \\
 & \multirow{2}{*}{Unseen Objects}
& Rag
& \greencheckmark
& \greencheckmark
& \greencheckmark
& \greencheckmark
& 1.0 / 1.0 / 4
 \\
 & & Rectangular Plate
& \greencheckmark
& \greencheckmark
& \greencheckmark
& \redxmark
& 0.0 / 1.0 / 3
  \\
 & \multirow{2}{*}{Lighting Changes}
& Dark
& \greencheckmark
& \greencheckmark
& \greencheckmark
& \greencheckmark
& 1.0 / 1.0 / 4
 \\
 & & Flickering
& \greencheckmark
& \greencheckmark
& \greencheckmark
& \greencheckmark
& 1.0 / 1.0 / 4
  \\
 & \multirow{2}{*}{Object Interference}
& Colorful Cubes 
& \greencheckmark
& \greencheckmark
& \greencheckmark
& \greencheckmark
& 1.0 / 1.0 / 4
 \\
 & & Multi-Plate
& \greencheckmark
& \greencheckmark
& \greencheckmark
& \redxmark
& 0.0 / 1.0 / 3
 
 \\
\midrule \midrule
\multirow{7}{*}{ACT}
& Normal
& Normal Setting
& \greencheckmark
& \greencheckmark
& \redxmark
& \greencheckmark
& 0.0 / 1.0 / 3
  \\
 & \multirow{2}{*}{Unseen Objects}
& Rag
& \greencheckmark
& \redxmark
& \redxmark
& \redxmark
& 0.0 / 0.0 / 1
 \\
 & & Rectangular Plate
& \greencheckmark
& \greencheckmark
& \redxmark
& \redxmark
& 0.0 / 1.0 / 2
  \\
 & \multirow{2}{*}{Lighting Changes}
& Dark
& \redxmark
& \redxmark
& \redxmark
& \redxmark
& 0.0 / 0.0 / 0
 \\
 & & Flickering
& \redxmark
& \redxmark
& \redxmark
& \redxmark
& 0.0 / 0.0 / 0
  \\
 & \multirow{2}{*}{Object Interference}
& Colorful Cubes 
& \greencheckmark
& \redxmark
& \redxmark
& \redxmark
& 0.0 / 0.0 / 1
 \\
 & & Multi-Plate
& \greencheckmark
& \redxmark
& \redxmark
& \redxmark
& 0.0 / 0.0 / 1
 
 \\
\midrule \midrule
\multirow{7}{*}{DP3}
& Normal
& Normal Setting
& \greencheckmark
& \greencheckmark
& \greencheckmark
& \greencheckmark
& 1.0 / 1.0 / 4
  \\
 & \multirow{2}{*}{Unseen Objects}
& Rag
& \greencheckmark
& \greencheckmark
& \greencheckmark
& \greencheckmark
& 1.0 / 1.0 / 4
 \\
 & & Rectangular Plate
& \greencheckmark
& \redxmark
& \redxmark
& \redxmark
& 0.0 / 0.0 / 1
  \\
 & \multirow{2}{*}{Lighting Changes}
& Dark
& \greencheckmark
& \greencheckmark
& \greencheckmark
& \greencheckmark
& 1.0 / 1.0 / 4
 \\
 & & Flickering
& \greencheckmark
& \greencheckmark
& \greencheckmark
& \greencheckmark
& 1.0 / 1.0 / 4
  \\
 & \multirow{2}{*}{Object Interference}
& Colorful Cubes 
& \redxmark
& \redxmark
& \redxmark
& \redxmark
& 0.0 / 0.0 / 0
 \\
 & & Multi-Plate
& \redxmark
& \redxmark
& \redxmark
& \redxmark
& 0.0 / 0.0 / 0
 
 \\
\midrule \midrule
\multirow{7}{*}{3D Diffuser Actor}
& Normal
& Normal Setting
& \greencheckmark
& \greencheckmark
& \redxmark
& \redxmark
& 0.0 / 1.0 / 2
  \\
 & \multirow{2}{*}{Unseen Objects}
& Rag
& \greencheckmark
& \greencheckmark
& \redxmark
& \redxmark
& 0.0 / 1.0 / 2
 \\
 & & Rectangular Plate
& \redxmark
& \greencheckmark
& \redxmark
& \redxmark
& 0.0 / 0.0 / 1
  \\
 & \multirow{2}{*}{Lighting Changes}
& Dark
& \greencheckmark
& \redxmark
& \redxmark
& \redxmark
& 0.0 / 0.0 / 1
 \\
 & & Flickering
& \greencheckmark
& \redxmark
& \redxmark
& \redxmark
& 0.0 / 0.0 / 1
  \\
 & \multirow{2}{*}{Object Interference}
& Colorful Cubes 
& \greencheckmark
& \greencheckmark
& \redxmark
& \redxmark
& 0.0 / 1.0 / 2
 \\
 & & Multi-Plate
& \greencheckmark
& \greencheckmark
& \redxmark
& \redxmark
& 0.0 / 1.0 / 2
 
 \\
\bottomrule
\end{tabular}}
\vspace{-5mm}
\end{table*}
\begin{table*}[htbp]
\centering
\caption{\textbf{Detailed generalization experiment results on Scan the Bottle task.}
}
\label{app_gen_scan}
\scalebox{0.6}{\begin{tabular}{ccccccc}
\toprule

Method & Case Index  & Pick the Bottle  & Pick the Scanner &  Scan  & Place Bottle in Box & SR (\%) $\uparrow$ /Loc-SR (\%) $\uparrow$ / Score $\uparrow$ \\
\midrule \midrule 
\multirow{3}{*}{PPI}&Normal Setting
& \greencheckmark
& \greencheckmark
& \greencheckmark
& \greencheckmark
& 1.0 / 1.0 / 4
  \\
&Box
& \greencheckmark
& \greencheckmark
& \greencheckmark
& \redxmark
& 1.0 / 1.0 / 3
 \\
&Yogurt Bottle
& \greencheckmark
& \greencheckmark
& \greencheckmark
& \redxmark
& 1.0 / 1.0 / 3
 \\
\midrule \midrule
\multirow{3}{*}{ACT}&Normal Setting
& \greencheckmark
& \greencheckmark
& \greencheckmark
& \greencheckmark
& 1.0 / 1.0 / 4
  \\
&Box
& \redxmark
& \greencheckmark
& \redxmark
& \redxmark
& 0.0 / 0.0 / 1
 \\
&Yogurt Bottle
& \redxmark
& \greencheckmark
& \redxmark
& \redxmark
& 0.0 / 0.0 / 1
 \\
\midrule \midrule
\multirow{3}{*}{DP3}&Normal Setting
& \greencheckmark
& \greencheckmark
& \greencheckmark
& \greencheckmark
& 1.0 / 1.0 / 4
  \\
&Box
& \redxmark
& \greencheckmark
& \redxmark
& \redxmark
& 0.0 / 0.0 / 1
 \\
&Yogurt Bottle
& \redxmark
& \greencheckmark
& \redxmark
& \redxmark
& 0.0 / 0.0 / 1

 \\
\midrule \midrule
\multirow{3}{*}{3D Diffuser Actor}&Normal Setting
& \greencheckmark
& \greencheckmark
& \greencheckmark
& \greencheckmark
& 1.0 / 1.0 / 4
  \\
&Box
& \redxmark
& \greencheckmark
& \redxmark
& \redxmark
& 0.0 / 0.0 / 1
 \\
&Yogurt Bottle
& \redxmark
& \greencheckmark
& \redxmark
& \redxmark
& 0.0 / 0.0 / 1
 \\
\bottomrule
\end{tabular}}
\vspace{-5mm}
\end{table*}

\end{document}